\documentclass{article}

\usepackage{arxiv}

\usepackage[utf8]{inputenc} 
\usepackage[T1]{fontenc}    
\usepackage{hyperref}       
\usepackage{url}            
\usepackage{booktabs}       
\usepackage{amsfonts}       
\usepackage{nicefrac}       
\usepackage{microtype}      
\usepackage{lipsum}
\usepackage{authblk}

\usepackage{float} 
\usepackage{times}
\usepackage{epsfig}
\usepackage{graphicx}
\usepackage{amsmath, amsfonts, amssymb}
\usepackage{bm}
\usepackage{algorithm}
\usepackage{algorithmicx}
\usepackage{comment}
\usepackage{fullpage}
\usepackage{tikz}
\usepackage{pifont}
\usepackage{amsthm}
\usepackage{subcaption}
\usepackage[square,sort,comma,numbers]{natbib}
\usepackage{enumitem}
\usepackage{hhline}
\usepackage{mathtools}
\usepackage{multirow}
\usepackage{algorithm}
\usepackage[noend]{algpseudocode}
\usepackage{booktabs}
\usepackage[export]{adjustbox}
\usepackage{url}
\usepackage{listings}
\usepackage{subcaption}
\usepackage{array}

\newcolumntype{C}[1]{>{\centering\let\newline\\\arraybackslash\hspace{0pt}}m{#1}}
\newcommand{\etal}{\textit{et al}.~}

\newcommand*{\vs}{\textit{vs. }}
\newcommand*{\wrt}{\textit{w.r.t. }}
\makeatletter
\newcommand*{\etc}{%
	\@ifnextchar{.}%
	{\textit{etc}}%
	{\textit{etc.}}%
}
\makeatother
\algtext*{EndWhile}
\algtext*{EndFor}
\algtext*{EndIf}
\algdef{SE}[DOWHILE]{Do}{doWhile}{\algorithmicdo}[1]{\algorithmicwhile\ #1}%

\makeatletter
\def\BState{\State\hskip-\ALG@thistlm}
\makeatother

\title{Prevention is Better than Cure: Handling Basis Collapse and Transparency in Dense Networks}

\author[1]{\textbf{Gurpreet Singh} \textsuperscript{\dag}}
\author[2]{\textbf{Soumyajit Gupta} \textsuperscript{\dag}}
\author[1]{\textbf{Clint N. Dawson}}
\affil[1]{Oden Institute for Computational Engineering and Sciences}
\affil[2]{Department of Computer Science}
\affil[ ]{University of Texas, Austin}
\affil[ ]{\texttt{\{gurpreet, smjtgupta\}@utexas.edu, clint@oden.utexas.edu}}

\begin{document}

\maketitle

{\let\thefootnote\relax\footnote{{\dag contributed equally to this work.}}}

\begin{abstract}
    Dense nets are an integral part of any classification and regression problem. Recently, these networks have found a new application as solvers for known representations in various domains. However, one crucial issue with dense nets is it's feature interpretation and lack of reproducibility over multiple training runs. In this work, we identify a \textit{basis collapse} issue as a primary cause and propose a modified loss function that circumvents this problem. We also provide a few general guidelines relating the choice of activations to loss surface roughness and appropriate scaling for designing low-weight dense nets. We demonstrate through carefully chosen numerical experiments that the basis collapse issue leads to the design of massively redundant networks. Our approach results in substantially concise nets, having $100 \times$ fewer parameters, while achieving a much lower $(10\times)$ MSE loss at scale than reported in prior works. Further, we show that the width of a dense net is acutely dependent on the feature complexity. This is in contrast to the dimension dependent width choice reported in prior theoretical works. To the best of our knowledge, this is the first time these issues and contradictions have been reported and experimentally verified. With our design guidelines we render transparency in terms of a low-weight network design. We share our codes for full reproducibility available at \url{https://github.com/smjtgupta/Dense\_Net\_Regress}.
\end{abstract}

\section{Introduction}

Dense network is primarily an Exploratory Data Analysis (EDA) tool that relies on statistical techniques for feature classification and regression over a data distribution. In principle, given infinite data, Universal Approximation Theorem (UAT) suggest that dense nets are powerful enough to represent any finite dimensional mapping between a set of inputs and corresponding outputs. However, in practice whether such a mapping can be efficiently and accurately learned is often difficult to achieve \cite{marcus2018deep}. One common concern with any gradient descent method over a non-convex surface is convergence to a local minimum resulting in a suboptimal solution.

The choice of activation/localization renders a directional nature to the dense network wherein the input data scales can be imbibed by the network. However, the same is not true for observed or true labels and therefore it is necessary to scale the observation data. In this work, we discuss some of the limitations of using tanh vs relu for regression. Supporting numerical experiments provide practical insights into the choice of activations and the consequent limitations. We discuss some of these choices for transparent design guidelines as per the definition proposed by Lipton \cite{lipton2018mythos}. 

We show that the network width and depth accommodate additive and multiplicative features which can be easily interpreted by a modeler working \wrt their application domain. We also identify a \textit{basis collapse} issue that manifests as loss drop rate stagnation or alternatively convergence to a local minimum, resulting in suboptimal solutions. Our simple numerical experiments demonstrate that this issue severely hampers the network learning capacity by redundantly accounting for a feature more than once. We also observed that this issue is much more pronounced for \textit{relu} activations due to the roughness of the loss surface where the fixed points act as strong attractors.
\begin{table}[h]
\centering
\caption{Performance Comparison for 2D}
\begin{tabular}{l|cccc}
\hline
Burgers' & Depth & Width & Params & MSE \\ \hline
PINN & $9$ & $20$ & $3441$ & $4.78e-3$ \\ 
Ours & $3$ & $4$ & $56$ & $1.28e-4$ \\ \hhline{=====}
Allen Cahn & Depth & Width & Params & MSE \\ \hline
PINN & $4$ & $200$ & $121401$ & $6.99e-3$ \\ 
Ours & $3$ & $6$ & $108$ & $3.64e-4$ \\\hline
\end{tabular}
\label{tab:early}
\end{table}
We propose an additional similarity loss along with the conventional Mean Squared Error (\textit{mse}) that prevents this issue by lifting the local minima corresponding to the collapse points on the loss surface. Experiments on multi-dimensional datasets show that this loss not only prevents \textit{basis collapse} but makes the network reproducible over multiple tranining runs. This is to say that the features identified across runs are similar for a network with fixed width and depth rendering us consistency. We show a snippet of our network's performance compared to a recently proposed representation-driven (Physics Informed Neural Network (PINN) \cite{raissi2018deep}) framework for bench-marking and verification purposes. \textbf{Table \ref{tab:early}} shows a comparison of our low-weight network design for a choice of two known PDE representations/2D datasets as follows:
\begin{align*}
    \tag{Burgers' Form}
    &u_{t} + uu_{x} = 0.01u_{xx} \label{eq:1dburger} \\
    \tag{Allen Cahn Form}
    &u_t + 5u^3 - 5u = u_{xx} \label{eq:1dallen}
\end{align*}
Note that while a PINN based framework for \textbf{Allen Cahn} form requires a large number of parameters $\mathbf{(10^6)}$ and achieves an \textit{mse} of $\mathbf{7e-3}$, our proposed net is much simpler in design and low weight having $(\mathbf{10^2)}$ parameters and still achieves \textit{mse} at $\mathbf{3.6e-4}$, suggesting that \textit{basis collapse} might be the primary issue leading to the choice of such a large network. Similar inference can be drawn for the \textbf{Burgers'} form. In the following text, we borrow notations from the established literature, where width $(w)$ and depth $(d)$ refers to the number of neurons in each layer and the number of hidden layers respectively, and the dimension of the input space is $n$. 

\section{Related Works}

Our work is mainly motivated by the findings and comments presented in \cite{li2018visualizing} and \cite{rendle2019difficulty}. Li \etal \cite{li2018visualizing} stated that when networks become sufficiently deep, neural loss landscapes quickly transition from being nearly convex to being highly chaotic. This transition from convex to chaotic behavior coincides with a dramatic drop in generalization error, and ultimately to a lack of trainability. Rendle \etal \cite{rendle2019difficulty} constructed simple two-dimensional examples where they showed that network has high chances of getting stuck in some bad local minima. Safran \etal \cite{safran2016quality} studied non-convex surface approximation through \textit{relu} functions and concluded that there theoretically exists an initialization and a basin where the global minima exists, which can be reached with high probability. Li \etal \cite{li2018visualizing} presents experimental manifestations of surface roughness generated using \textit{relu} network. They show, by visualizing lower dimensional manifolds, that introducing skip connections makes the high dimensional surface smoother. PINN \cite{raissi2018deep,raissi2019physics} framework has showcased dense nets as solvers for functional representations. They present a number of numerical experiments with known forms of PDEs with the aim to approximate the functional form of the solution using a dense net. Prior theoretical works that might assist in transparent network designs are in the form of Universal Approximation Theorems (UAT) where a strict dependency of network parameters on the input data dimension $(n)$ is shown. While Hanin \etal \cite{hanin2017approximating} states that for an arbitrary depth, a dense net with \textit{relu} activation can approximate any continuous function in a bounded $L_1$ space if width is $w=n+1$, Lu \etal \cite{lu2017expressive} make a similar claim but with width $w=n+4$.

\section{Motivation} \label{sec:motivation}

We motivate this section by pre-emptively showing a low weight network $(w,d)=(4,4)$ that works equally well for two problems of varying complexity as described by the equations below. We do away with the prescription of boundary and initial conditions and focus only on the feature complexity of the obtained solution to motivate our work.\\ 
\noindent \textbf{Form 1: Burgers' equation for shock formation}
\begin{align*}
    &u_{t} + uu_{x} + uu_{y} = 0.01 (u_{xx} + u_{yy})
\end{align*}
\noindent \textbf{Form 2: General viscous Burgers' equation}
\begin{align*}
    &u_{t} + uu_{x} + vu_{y} = 0.01 (u_{xx} + u_{yy})\\
    &v_{t} + uv_{x} + vv_{y} = 0.01 (v_{xx} + v_{yy})
\end{align*}
Although at a glance the second form might look more hefty (two equations and two unknowns), we show that both are sufficiently resolved with the low weight network described above. The only difference is the number of outputs: one (only $u$) for the first form and two (both $u,v$) for the second form. 
\begin{figure}[h]
\centering
    \begin{subfigure}[b]{0.49\linewidth}
		\centering
		\includegraphics[trim={2cm 0 4.5cm 0},clip,width=\linewidth]{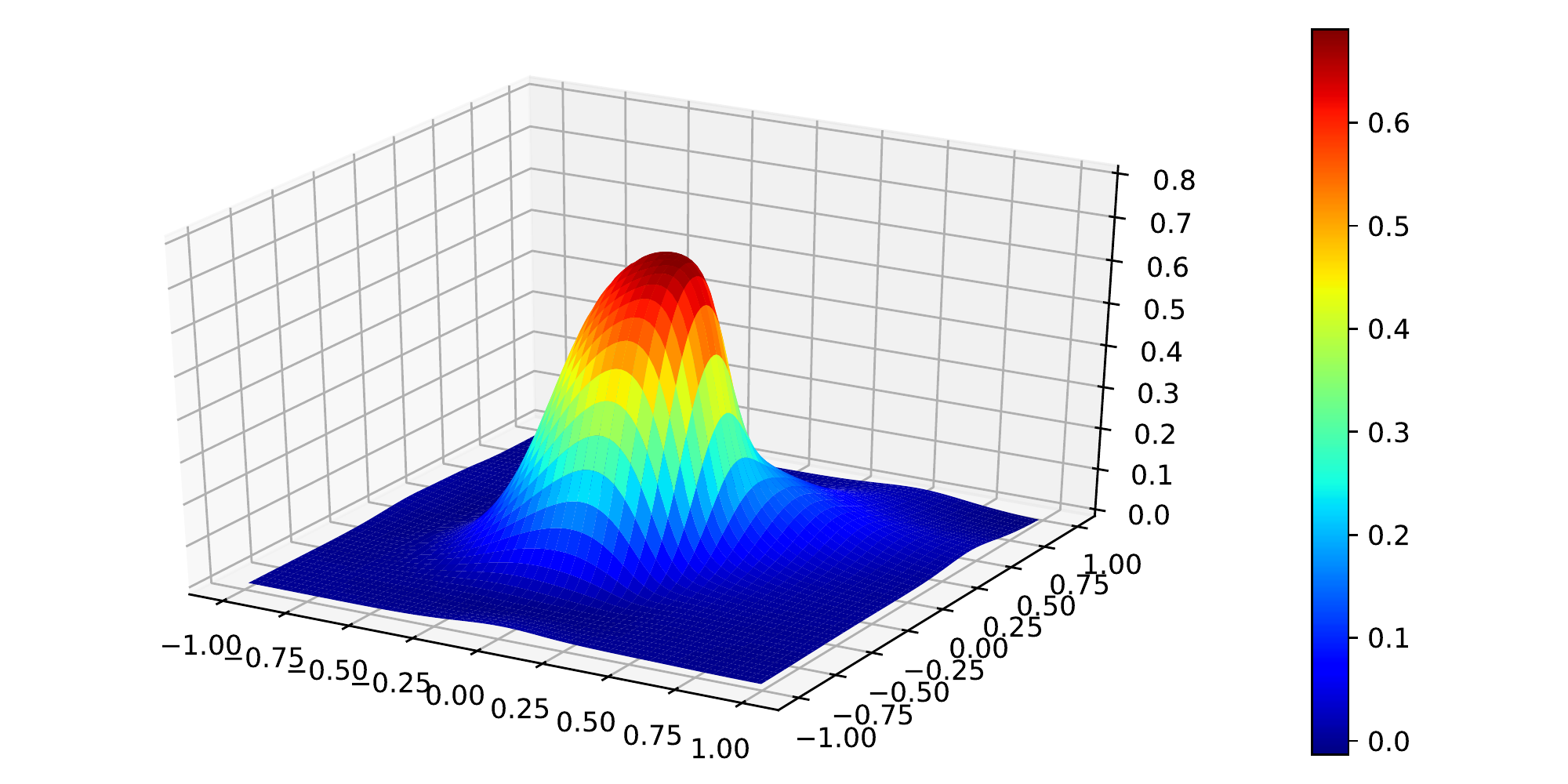}
		\caption{Surface at t = 0.0}
	\end{subfigure}
	\begin{subfigure}[b]{0.49\linewidth}
		\centering
		\includegraphics[trim={2cm 0 4.5cm 0},clip,width=\linewidth]{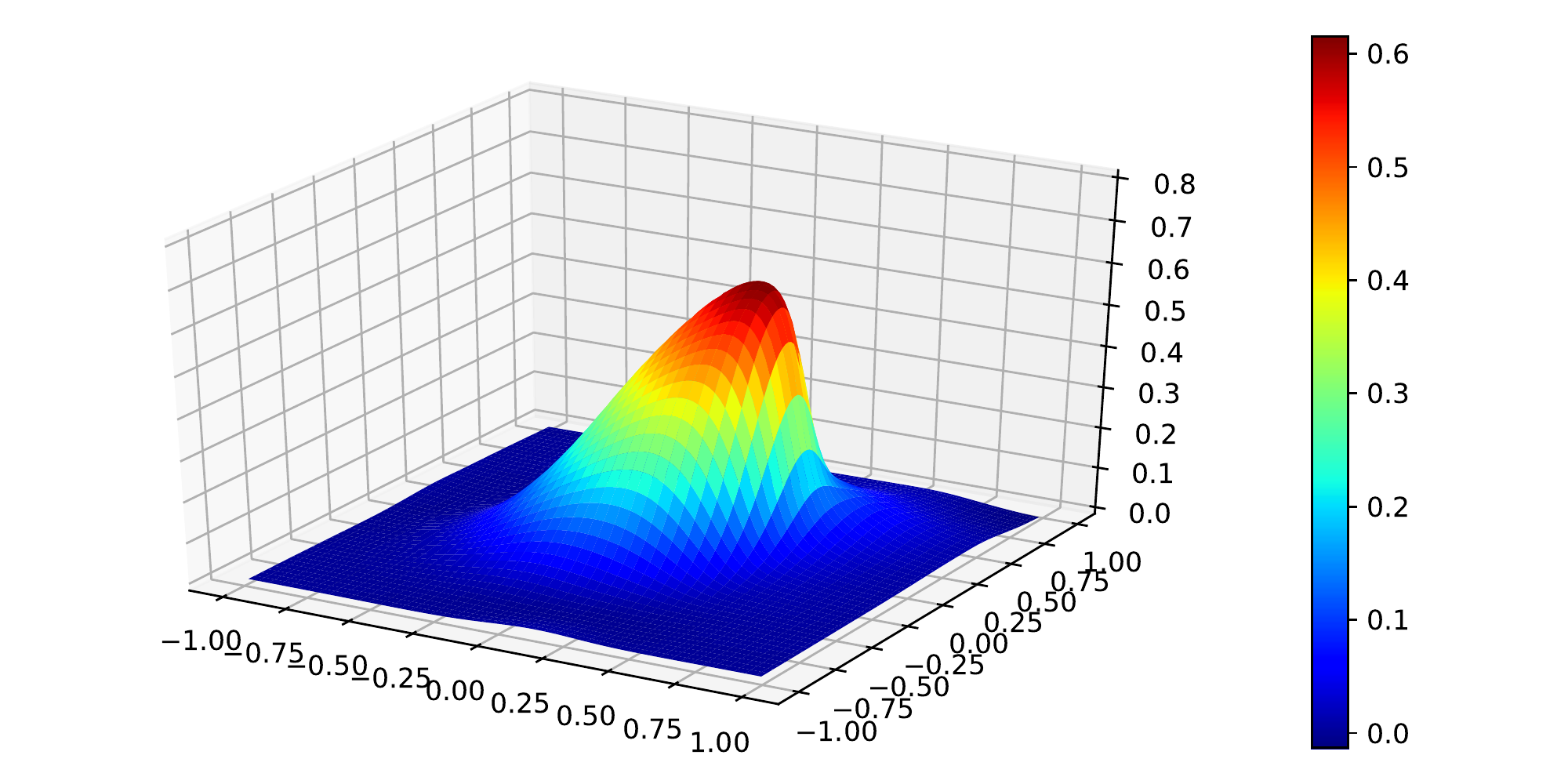}
		\caption{Surface at t = 0.5}
	\end{subfigure}
    \caption{Solving Burgers equation (shock formation) using our approach. \textit{mse} loss: $2.18e-4$ at $30$ epochs with $lr=0.2$}
    \label{fig:bg_2d_f1}
\end{figure}

\begin{table}[h]
    \centering
    \caption{Comparison of Burgers 3D. `-' indicate redundant values because the network did not converge to the solution}
    \begin{tabular}{lcccc}
        Arch & Depth & Width & Params & MSE \\ \hline
        PINN & $9$ & $20$ & $3482$ & $-$ \\
        Ours & $4$ & $4$ & $60$ & $2.18e-4$ \\ \hline
    \end{tabular}
    \label{tab:bg_intro}
\end{table}

\begin{figure}[h]
\centering
    \begin{subfigure}[b]{0.49\linewidth}
		\centering
		\includegraphics[trim={0.5cm 0 0.5cm 0},clip,width=\linewidth]{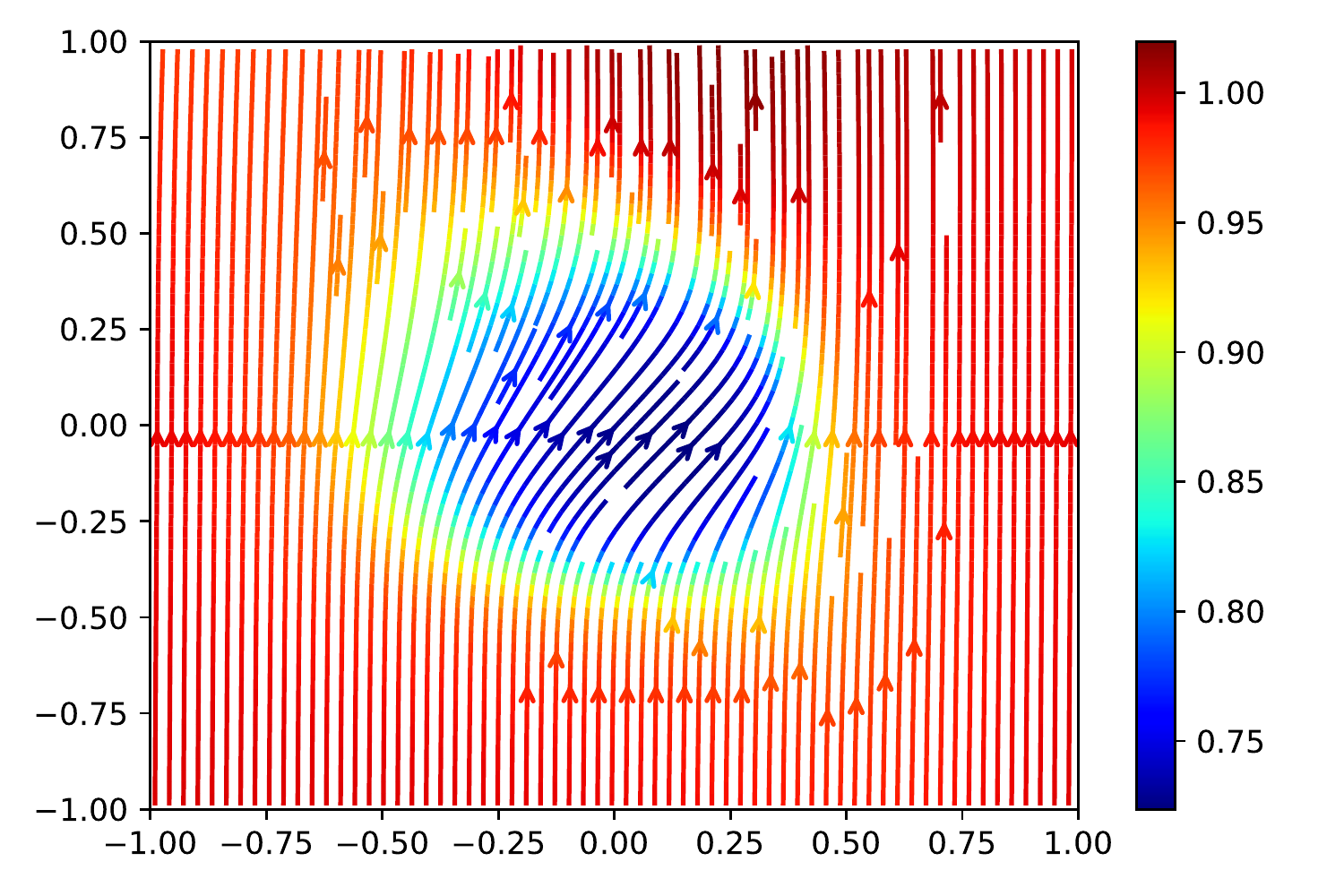}
		\caption{Stream Plot at t = 0.0}
	\end{subfigure}
	\begin{subfigure}[b]{0.49\linewidth}
		\centering
		\includegraphics[trim={0.5cm 0 0.5cm 0},clip,width=\linewidth]{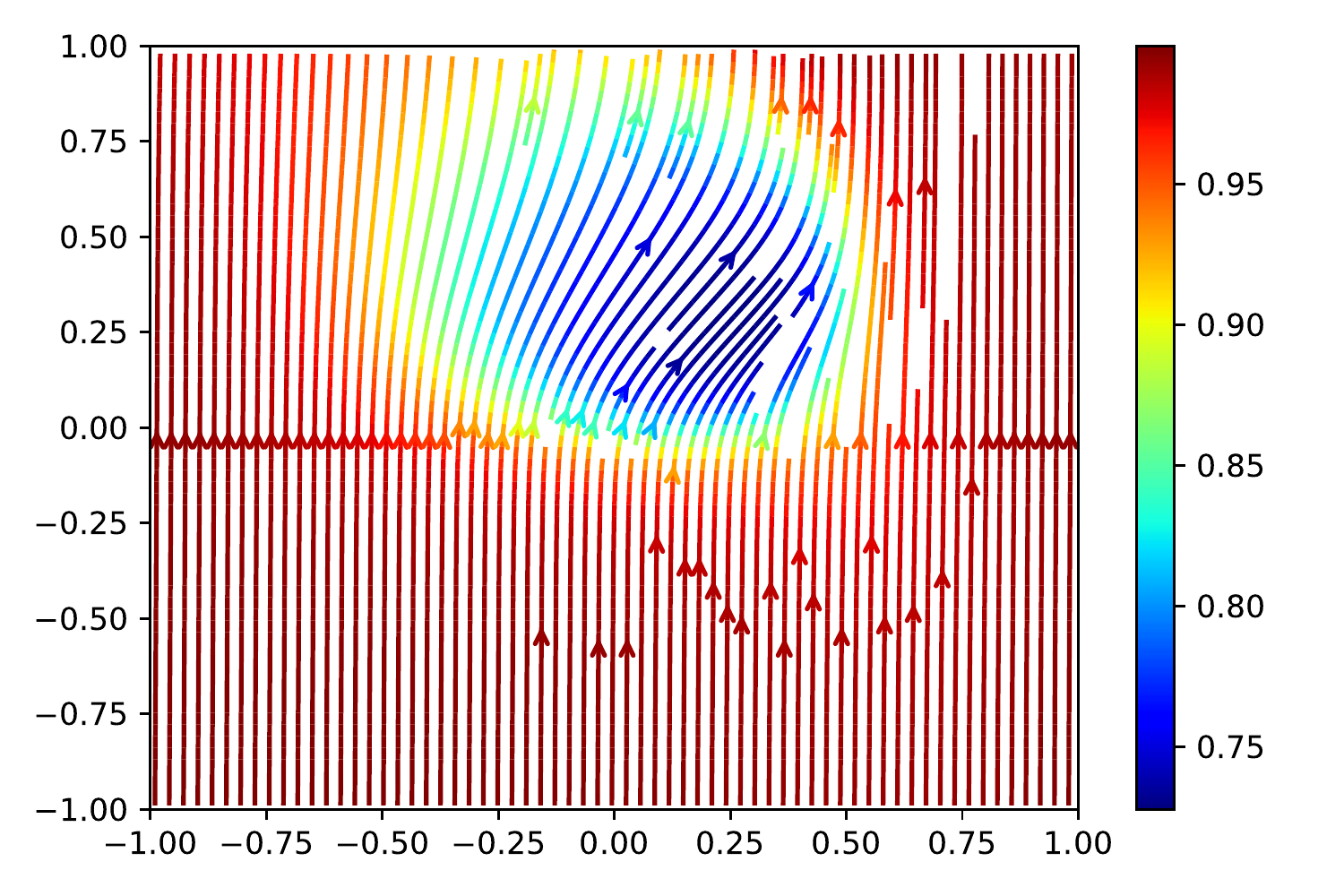}
		\caption{Stream Plot at t = 0.5}
	\end{subfigure}
    \caption{Solving a general viscous Burgers/ flow using our approach. \textit{mse} loss: $5.27e-4$ at $30$ epochs with $lr=0.2$}
    \label{fig:bg_2d_f2}
\end{figure}

\textbf{Figs. \ref{fig:bg_2d_f1}} and \textbf{\ref{fig:bg_2d_f2}} show the scalar $(u)$ and vector $(u,v)$ solution for the two forms at time slices $t=0$ and $t=0.5$. In Fig. \ref{fig:bg_2d_f2}, the arrows represent the vector direction and the colors the magnitude of the vector. Both solution were achieved with a \textit{mse} loss at scale $e-4$. We later show why the scalar (one output) and vector (two outputs) datasets in two spatial and one temporal dimensions require the same network design due to the presence of additive and multiplicative features resolved in the network width and depth, respectively. \textbf{Table \ref{tab:bg_intro}} shows a performance comparison between the recently introduced PINN framework \cite{raissi2018deep,raissi2019physics} and our design once a few challenges are sufficiently addressed.

We now pose the following questions motivated by experiences gathered from training dense networks to improve network transparency and interpretability:
\begin{enumerate}[leftmargin=*]
    \item Can a trained network be used for predicting outside the data domain it was trained on?
    \item How does the choice of activation/localization function impact feature learning?
    \item How do dense nets account for features in width and depth?
    \item A number of UATs indicate dependence of width and depth on data dimensionality. What are the practical implementation challenges in deploying these findings?
    \item With the network design held in place, why do we need to prescribe multiple runs to achieve a good representation of the features or obtain a desirable loss value?
    \item What is the difference between data driven and recently introduced representation driven network learning?
\end{enumerate}

In the following sections, we build upon each of these questions to identify a \textit{basis collapse} issue that once sufficiently addressed, renders consistency between multiple training runs. Our main contributions are:
\begin{enumerate}
    \item Design guidelines to develop low weights (computationally inexpensive) networks that adequately represent the underlying features in an effort to improve transparency.
    \item Modified loss function that prevents basis collapse to achieve consistency across runs.
\end{enumerate}

\section{Setup and training}
All experiments were done on a setup with Nvidia 2060 RTX Super 8GB GPU, Intel Core i7-9700F 3.0GHz 8-core CPU and 16GB DDR4 memory. We use Keras \cite{chollet2015} library running on top of Tensorflow 1.14 backend with Python 3.5 and \textit{AdaGrad} \cite{duchi2011adaptive} as optimizer.

\section{Limitations of Activation Functions}

In this section, we point out a few limitations that arise from the choice of activation/localization functions. We restrict our choice of activation functions to \textit{tanh} and \textit{relu} since they are diametrically opposite in terms of smoothness of the learned features. \textit{tanh} renders its smoothness (regularizes) to the learned basis whereas \textit{relu} preserves the maximum smoothness of the learned basis up till it's localization. The output of any dense network, anything from shallow to deep, is guided by this choice. To illustrate their limitations, we compared the two in terms of approximating a function and consequent impact on the solution space.

\begin{figure}[h]
    \centering
    \begin{subfigure}[b]{0.49\linewidth}
		\centering
		\includegraphics[width=\linewidth]{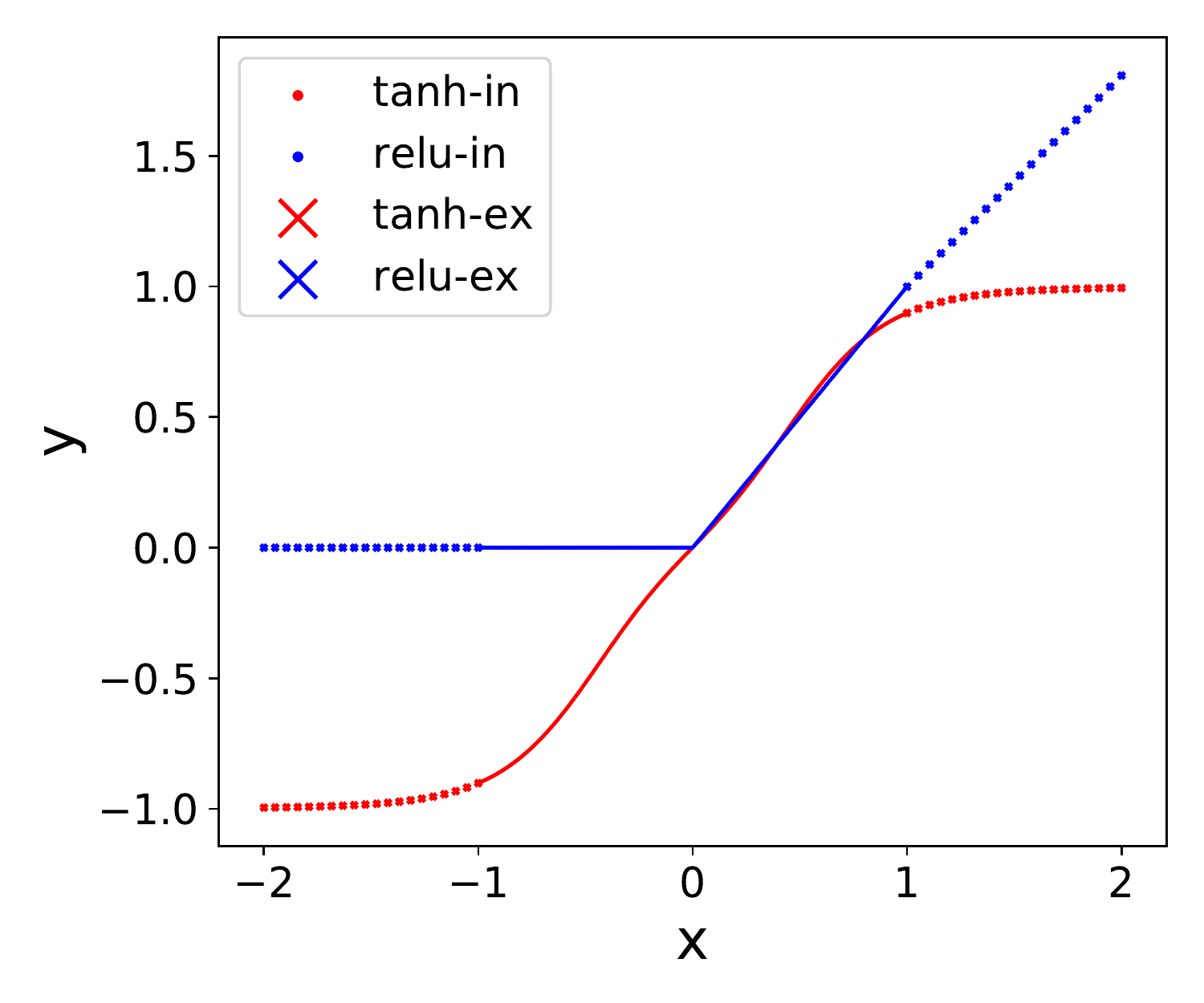}
		\caption{Predicting $y=x$}
	\end{subfigure}
	\begin{subfigure}[b]{0.49\linewidth}
		\centering
		\includegraphics[width=\linewidth]{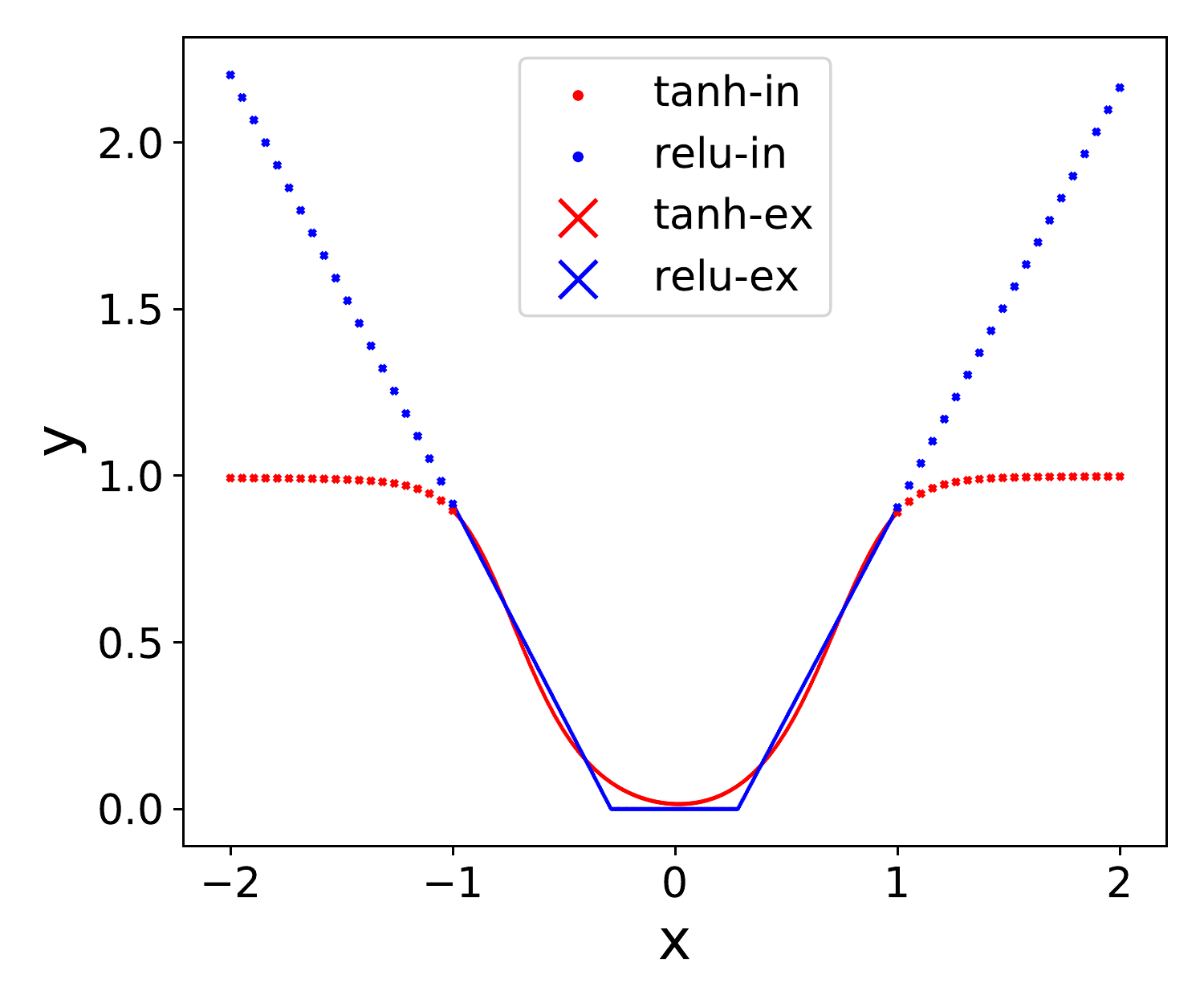}
		\caption{Predicting $y=x^2$}
	\end{subfigure}
    \caption{$y=x,x^2$ at $x=[-1,1]$. Prediction inside (\{\textit{tanh}, \textit{relu}\}-in) and outside (\{\textit{tanh}, \textit{relu}\}-ex) the domain. Interpolation with \textit{tanh} is smoother, while Extrapolation deviates and flattens out. \textit{relu} performs better linearly within a certain range.}
    \label{fig:extrap}
\end{figure}

\subsection{Function Approximation} \label{sec:extrap}

We consider fitting a simple $(w,d)=(3,1)$ dense-net with \textit{relu} and \textit{tanh} activations for two functions $y=f(x)=x,x^2$. The network is trained on $500$ samples drawn uniformly within $[-1,1]$ and predictions are made on both interpolated $[-1,1]$ and extrapolated points $[-2,-1],[1,2]$. In Fig. \ref{fig:extrap} we see that while \textit{tanh} cannot predict anything beyond its learnt range, \textit{relu} has the capability of extrapolating points linearly. We would like to emphasize that although the output of the interpolated points looks like approximating a parabola, it is clearly evident that beyond the trained domain, extrapolation deviates away from the true function by inheriting smoothness of the activation function. Thus it predicts $y=1.0, 2.3$ instead of $y=4.0$ at $x=\pm 2$ in \textbf{Fig. \ref{fig:extrap}} for \textit{tanh, relu} respectively. 

\begin{figure}[h]
    \centering
    \begin{subfigure}[b]{0.49\linewidth}
		\centering
		\includegraphics[width=\linewidth]{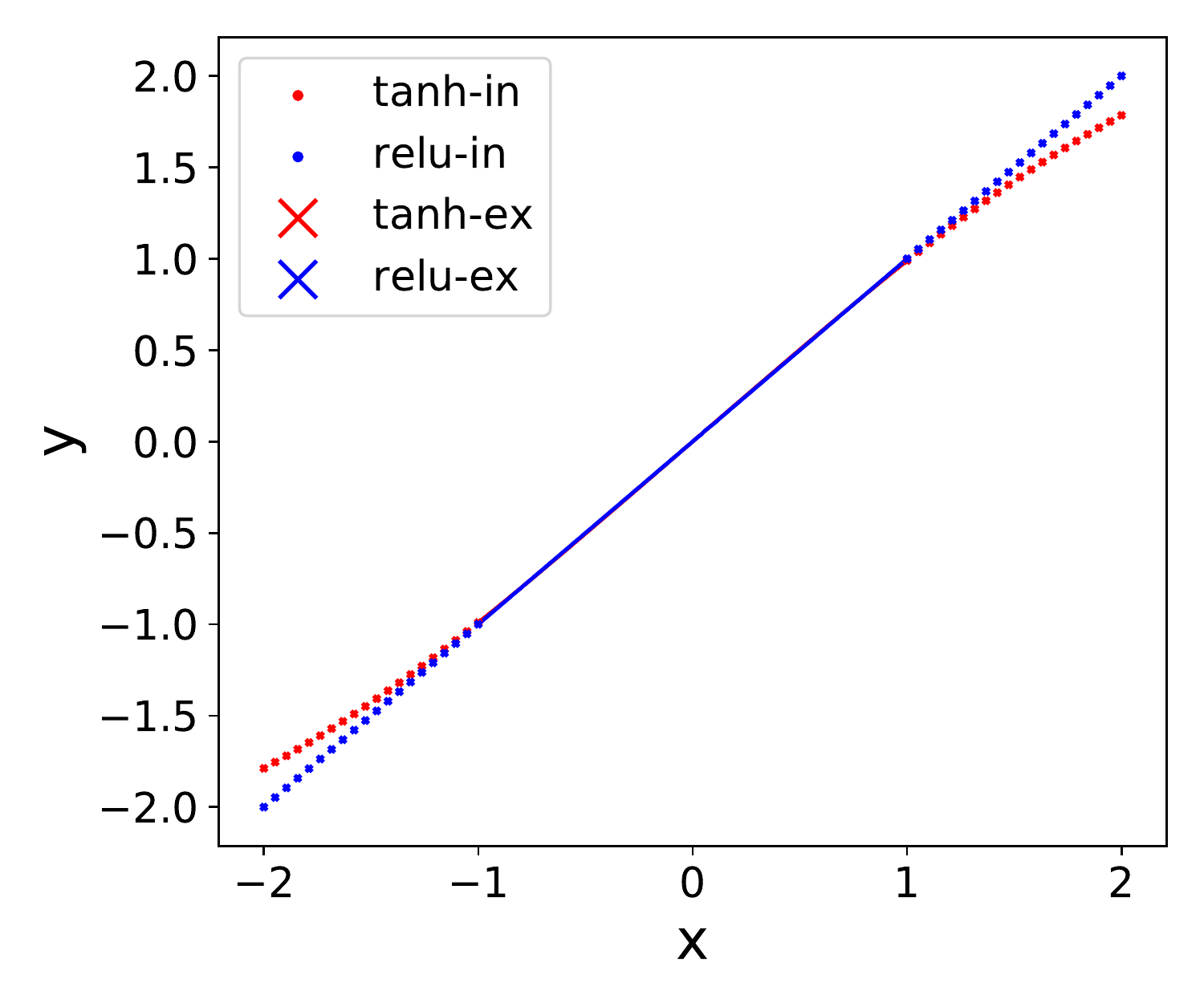}
		\caption{Predicting $y=x$}
	\end{subfigure}
	\begin{subfigure}[b]{0.49\linewidth}
		\centering
		\includegraphics[width=\linewidth]{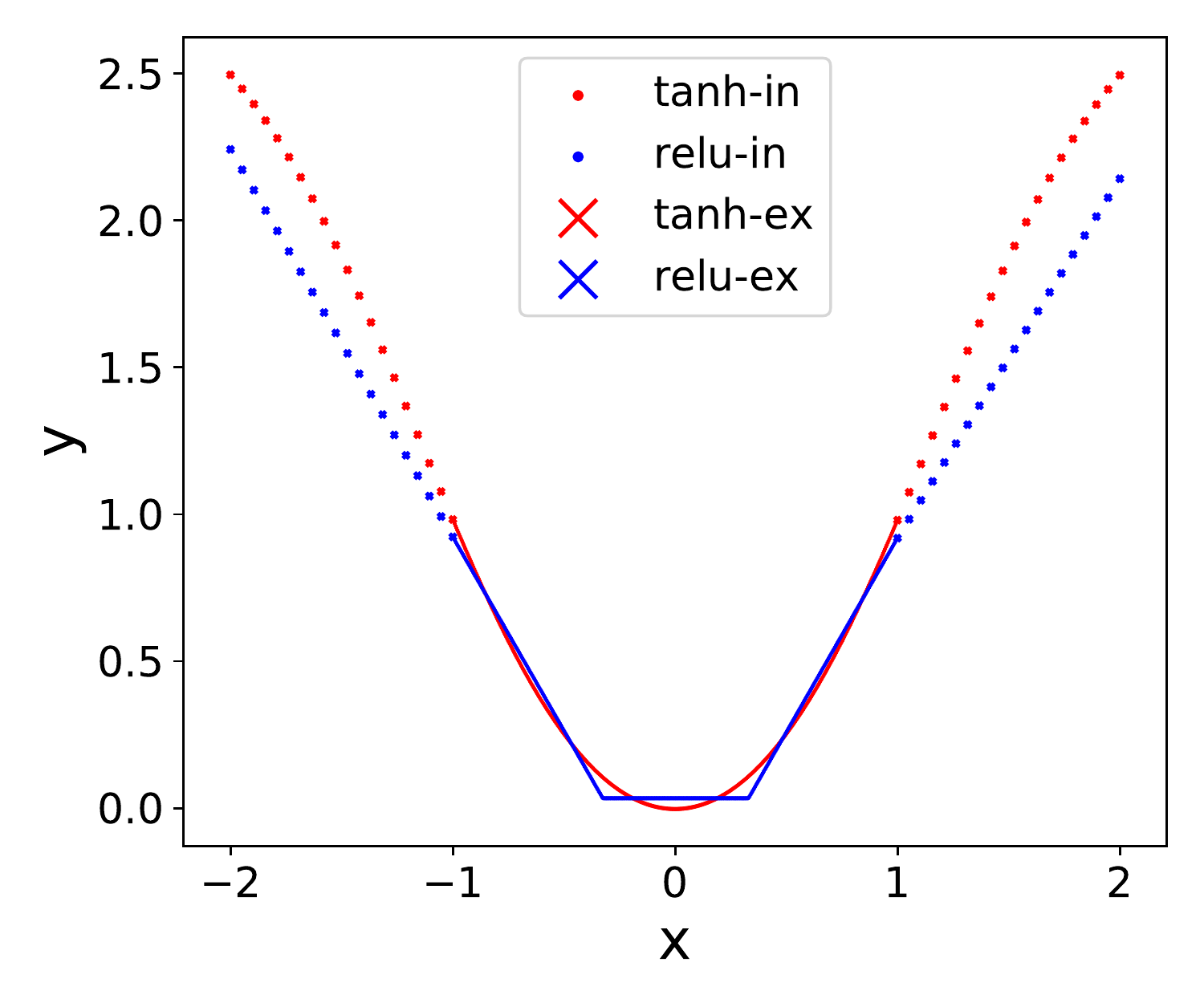}
		\caption{Predicting $y=x^2$}
	\end{subfigure}
    \caption{$y=x,x^2$ at $x=[-1,1]$. Prediction inside and outside the trained domain. Apart from the output layer accounting for the data scale now, the extrapolation capability is still null.}
    \label{fig:extrap-lin}
\end{figure}

To show the importance of the output layer during dense net regression, we keep the same $d=1$ hidden layer setting as above, but let the output layer to have \textit{linear} activation in \textbf{Fig. \ref{fig:extrap-lin}} \footnote{This is the standard setup used in dense net regression, wherein the hidden layers have user-chosen activation, while the final output layer has \textit{linear} activation and zero bias. All subsequent networks follow such a design.}. During back-propagation of errors, the entire responsibility of approximating the scales falls on this output layer. Still, the extrapolation problem persists as the network predicts $y=2.25,2.5$ instead of $y=4.0$ at $x=\pm 2.0$ for \textit{relu} and \textit{tanh} respectively. Thus, even with \textit{linear} activation at the output, the network was barely enough to fit the data, rather than approximating the function $y=x^2$. It is our understanding that conclusions using extrapolation of the learned representation must be, at best, approached with extreme caution irrespective of the data dimensionality. This hold true for all the learned representations for $>1$ dimensional datasets considered in this work and can be trivially verified.

\subsection{Hidden Parameter Landscape and Activation Functions} \label{sec:act}

\begin{figure}[h]
    \centering
    \begin{subfigure}[b]{0.49\linewidth}
		\centering
		\includegraphics[width=\linewidth]{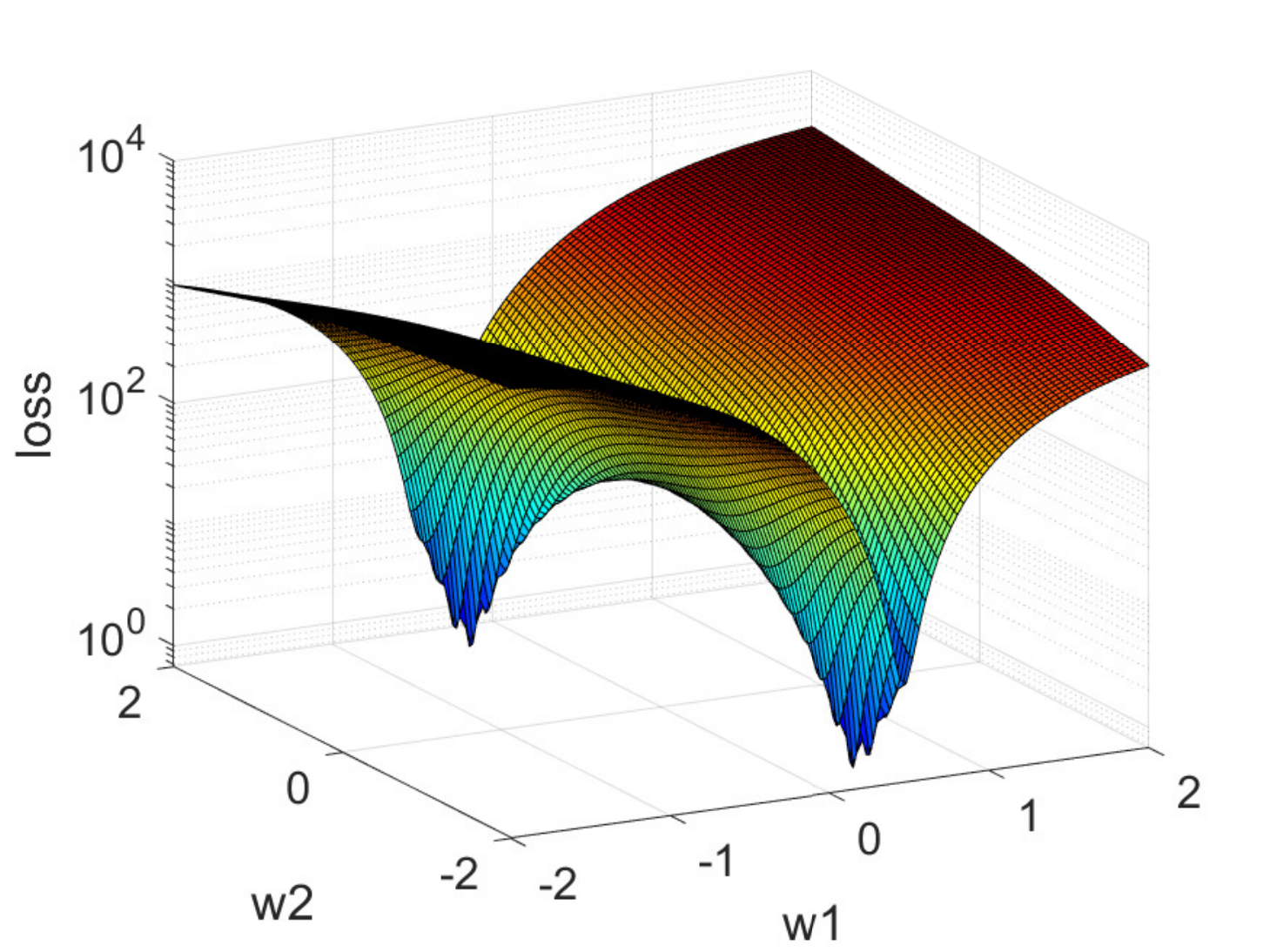}
		\caption{\textit{tanh} surface}
	\end{subfigure}
	\begin{subfigure}[b]{0.49\linewidth}
		\centering
		\includegraphics[width=\linewidth]{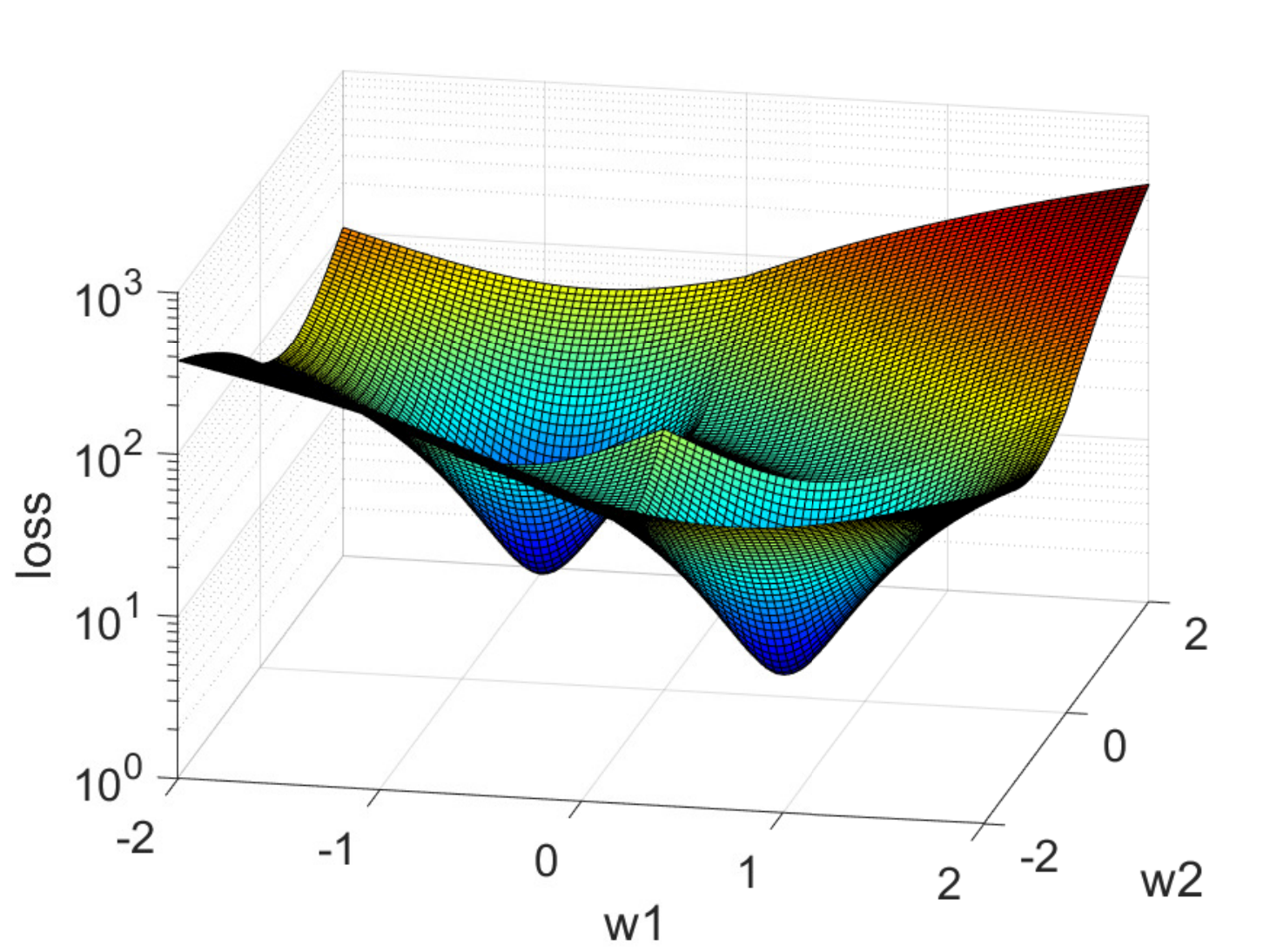}
		\caption{\textit{relu} surface}
	\end{subfigure}
    \caption{Loss induced hidden parameter landscape for \textit{tanh} and \textit{relu} activations. \textit{tanh} has comparably lower loss \wrt \textit{relu}. $\textrm{w1},\textrm{w2}$ are the weights of the first hidden layer.}
    \label{fig:surface}
\end{figure}

To illustrate the dependency of the solution surface on the choice of activation/localization function, we plot the hidden parameter surface of the dense net for both \textit{relu} and \textit{tanh} in \textbf{Fig. \ref{fig:surface}}. It is important to note that the fixed points of the Fig. \ref{fig:surface} (b) are strong attractors when compared to the fixed points of Fig. \ref{fig:surface} (a). Notice that the descent direction at any point near this attractor is approximately the same and points to the minimum in the vicinity. This in turn makes a network with \textit{relu} activations more prone to converging to a local minima making stochastic gradient descent escape less probable. 

Note that the surfaces shown in this figure are true surfaces considering all the sample points. A stochastic gradient descent approximates this surface by taking into account a batch of points to approximate the descent direction. Nevertheless, the nature of the difficulty remains unchanged. For this reason, all the subsequent sections use a \textit{tanh} activation for the hidden layers. Also note that the surface roughness in Fig. \ref{fig:surface} (a) is a visualization artifact and the true surface is smooth.

\subsection{Scale Independent Representation of Functions} \label{ssec:scale_independent}

\begin{figure}[h]
    \centering
    \begin{subfigure}[b]{0.49\linewidth}
		\centering
		\includegraphics[width=\linewidth]{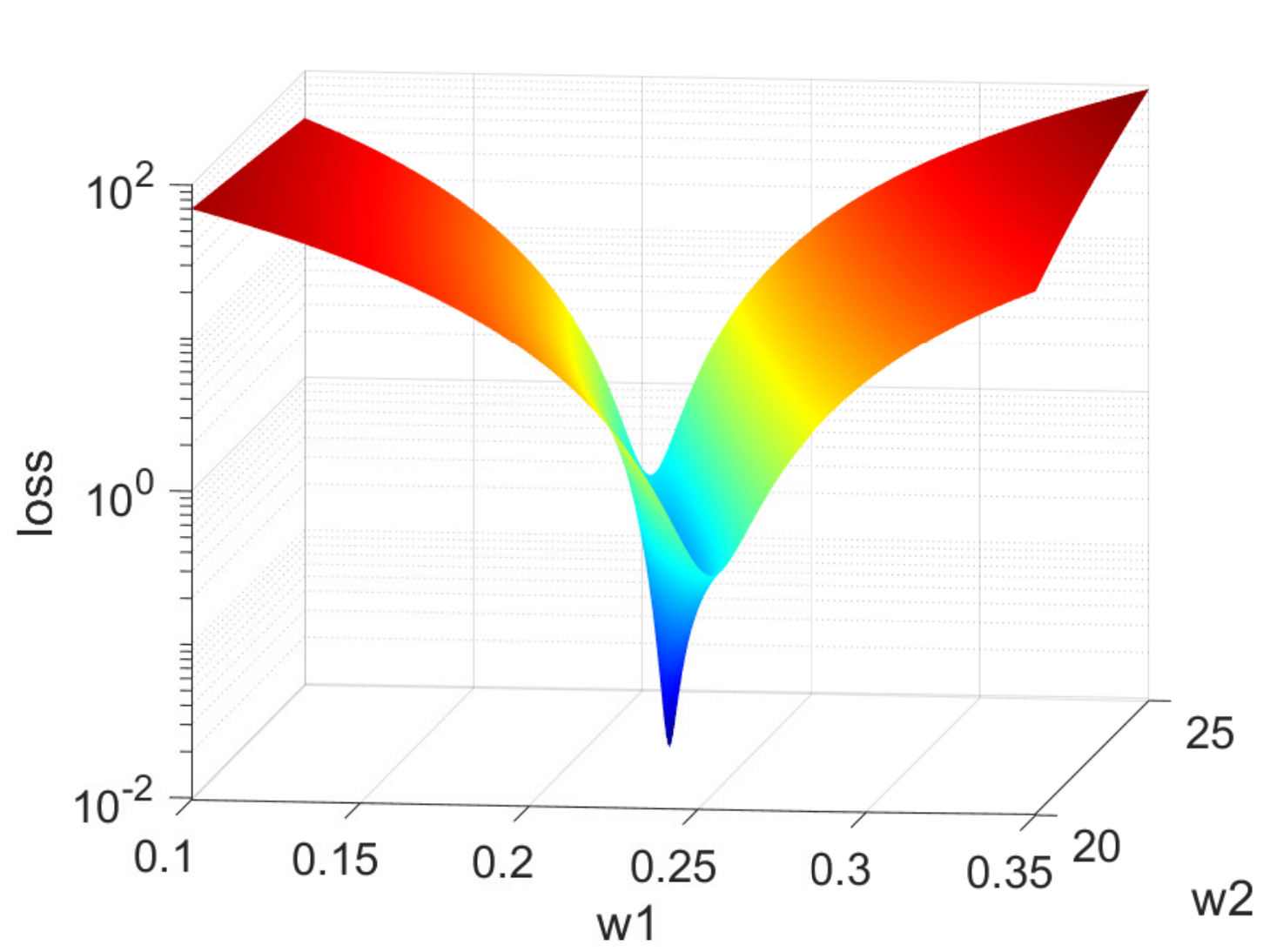}
		\caption{Surface without scaling}
	\end{subfigure}
	\begin{subfigure}[b]{0.49\linewidth}
		\centering
		\includegraphics[width=\linewidth]{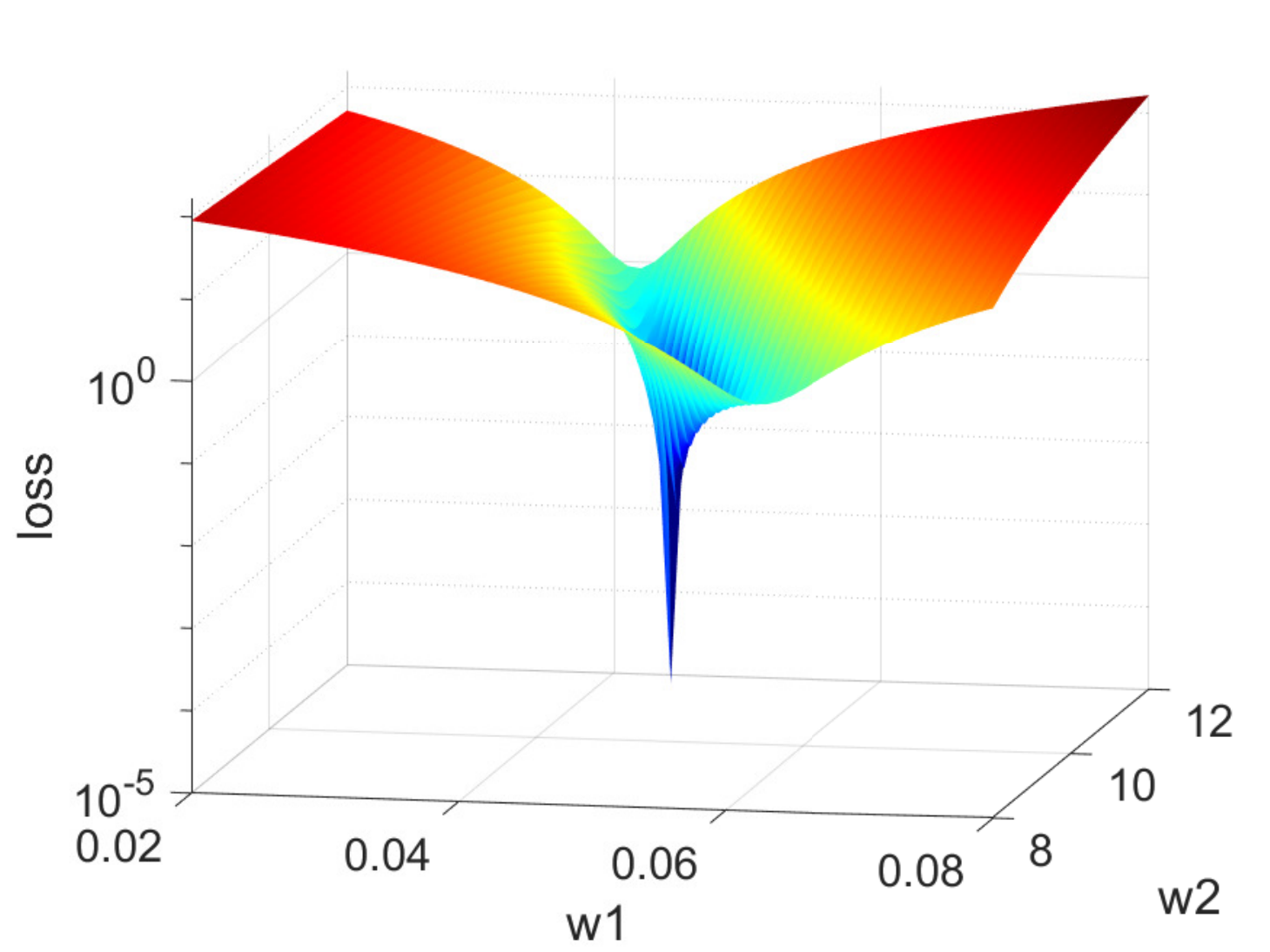}
		\caption{Surface with scaling}
	\end{subfigure}
    \caption{Loss surface for \textit{relu} and \textit{tanh}. $\textrm{w1},\textrm{w2}$ are the weights of the linear basis at $d=1,2$ respectively. The two loss functions are appropriately scaled for comparison.}
    \label{fig:surf_scal}
\end{figure}

We now point out a problem that arises if the labels are not appropriately scaled corresponding to the choice of activation functions. \textbf{Fig. \ref{fig:surf_scal}} shows the hidden parameter surface corresponding to a width 1 and depth 2 dense network without and with scaling the labels. This trivial network uses \textit{tanh} activation for the hidden layer. The dataset is a straight line $y=x$ as shown in Fig. \ref{fig:extrap-lin} (a) with $x,y \in [-2,2]$. In this setting, one would expect that this trivial network should be able to approximate the straight line without scaling the labels since the last linear layer can accommodate this scaling factor. However, as seen in Fig. \ref{fig:surf_scal} this argument does not hold. Fig. \ref{fig:surf_scal} (b) shows that the network can reach a lower values of loss if the labels are scaled between $(-1,1)$ in tandem with the choice of \textit{tanh} activation function with a range of $[-1,1]$. 

\begin{figure}[h]
    \centering
    \begin{subfigure}[b]{0.49\linewidth}
		\centering
		\includegraphics[width=\linewidth]{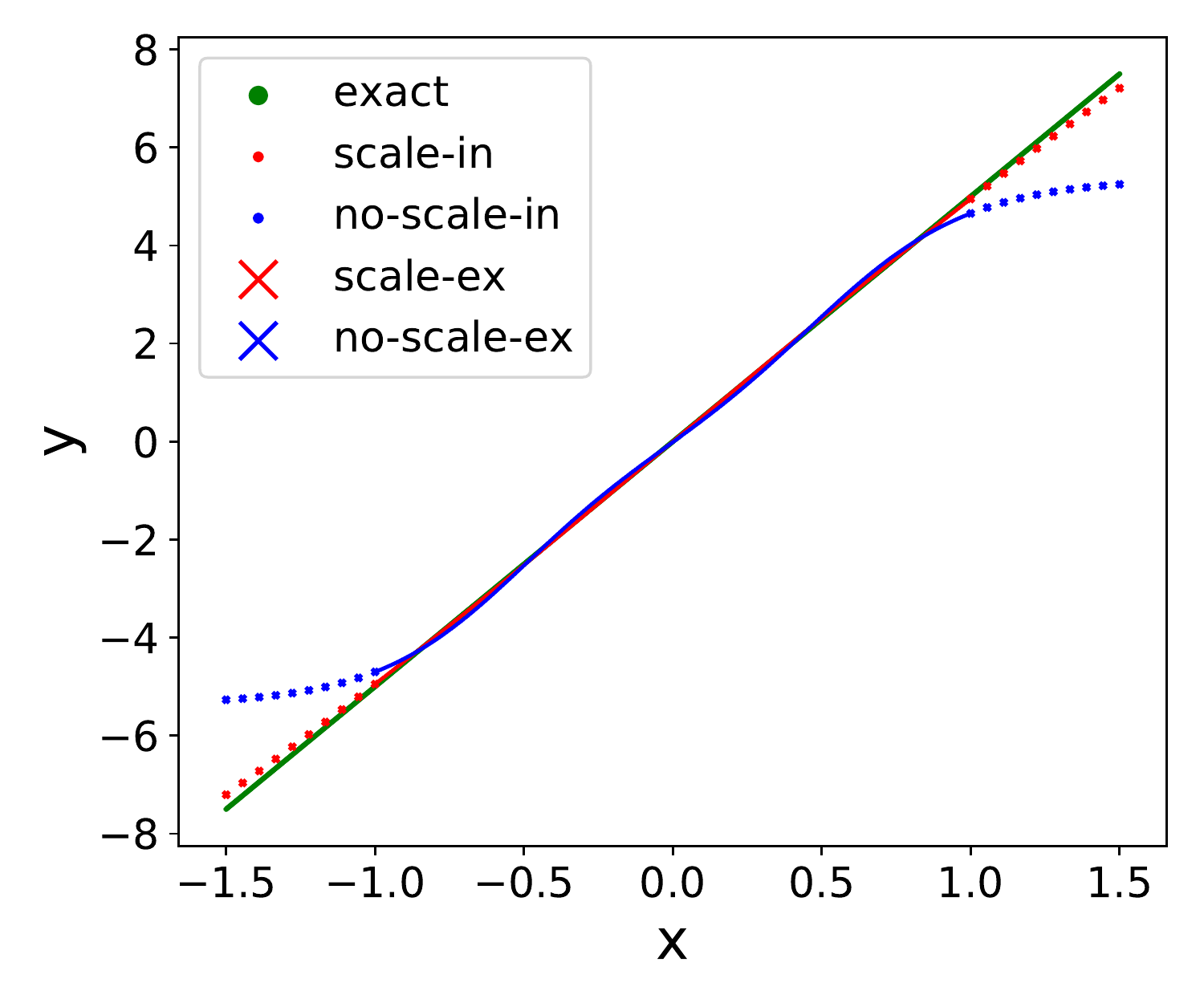}
		\caption{Predicting $y=5x$}
	\end{subfigure}
	\begin{subfigure}[b]{0.49\linewidth}
		\centering
		\includegraphics[width=\linewidth]{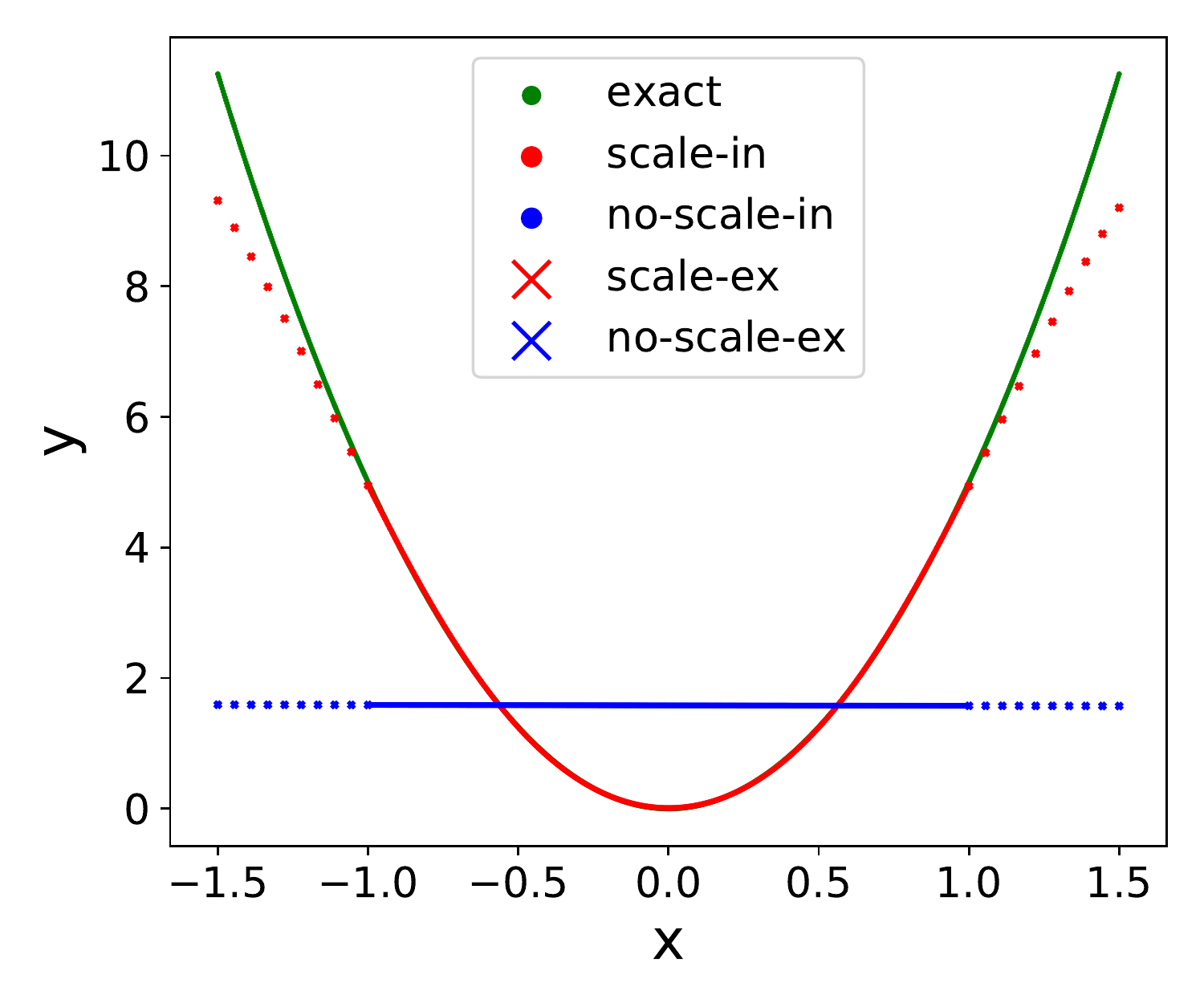}
		\caption{Predicting $y=5x^2$}
	\end{subfigure}
    \caption{Effect of scaling output labels on the network learning capacity.  Prediction inside (\{$no-scale$, $scale$\}-in) and outside \{$no-scale$, $scale$\}-ex) the trained domain are shown as solid lines and crosses, respectively.}
    \label{fig:fit_scal}
\end{figure}

We would like to point out that any other choice of scaling that does not conform with the range of the activation function, leads to lifting of minima on the hidden parameter induced loss surface in Fig. \ref{fig:surf_scal}. This is true even when the last layer (linear activation) is chosen specifically to accommodate scaling in the labels. Note that the $mse$ loss functions for the unscaled and scaled loss landscapes are $||u_{pred}-u_{true}||_{2}$ and $||s\times (u_{pred}-u_{true})||_{2}$, respectively. Here, $s$ is the scaling factor, to make the two loss surfaces comparable.\textbf{ Fig. \ref{fig:fit_scal} (a)} shows the fit to a linear dataset ($y = 5x$) using a $(w,d)=(1,1)$ network with (red) and without scaling (blue) making it visually evident as to which one performs better. \textbf{Fig. \ref{fig:fit_scal} (b) }shows a similar comparison for a quadratic function ($y = 5x^2$) with a network of $(w,d)=(2,1)$. For both the cases the data is in $x \in [-1,1]$. Fig. \ref{fig:fit_scal} (b) shows that in the absence of scaling, the network converges to the mean value of the input data, disregarding all features of the parabola, while the scaled version fits well.

We now discuss the manifestations of this problem in designing a network wherein we do not have the luxury of visualizing the fit adequately. The choice of not scaling the labels falsely gives the impression that the network width needs to be redundantly increased in order to fit the data better. The biases in this redundant width network will now compensate for this ill made choice while the weights remain similar consequently increasing the number of parameters. We call this \textit{additive redundancy} since the difference only appears in the biases while the weights remain similar curtailing network interpretability.

Computationally, not scaling the output labels appropriately results in an ill-conditioned minimization problem. This can be seen easily with the trivial linear dataset as in Fig. \ref{fig:extrap-lin} (a). If $y \in [-a,a]$, the bias in layer 1 (\textit{tanh} activation) must learn a quantity close to $b_{1} \approx 1/a$ while the bias in layer 2 (\textit{tanh} activation) must compensate by learning $b_{2} \approx a$. As $a$ increases the ratio of the two biases increases as $b_{2}/b_{1} \approx a^{2}$, making it difficult for SGD \citep{bottou2010large} to find the minimum. This issue is trivially resolved if the the output labels are scaled in accordance with the choice of activation function range. 

A question that one might immediately ask: Does $\textrm{w2}=10$ weight in layer 2 in Fig. \ref{fig:surf_scal} (b) not create conditioning issues? Yes it does, however now increasing the depth of the network will reduce this quantity to less than 1 for scaled output labels. The reason we abide by the above choices is to say that in the width we want to learn additive features whereas, the depth provides multiplicity to each of these additive features. 
These design choices now let us interpret any vanilla dense network quite easily. The multiplicity interpretation is further expanded upon by considering symmetric and asymmetric functions in the following section. The insights gained from these trivial examples extend to complicated cases where we design dense networks that are both low weights and transparent. 
 
In all of the following numerical experiments, we scale the output labels to circumvent the obstacles identified here. As will be shown later, if the objective is to solve a system of Ordinary or Partial Differential Equations (ODE, PDE) with densely connected networks, this implies working with scaled, dimensionless forms of the equations to get a low-weights and transparent network. These considerations make the dense network indifferent to the choice of the problem being addressed rendering a common basis for network interpretability. 
 
\section{Interpreting Width and Depth}

\begin{figure}[h]
    \centering
	\includegraphics[width=\linewidth]{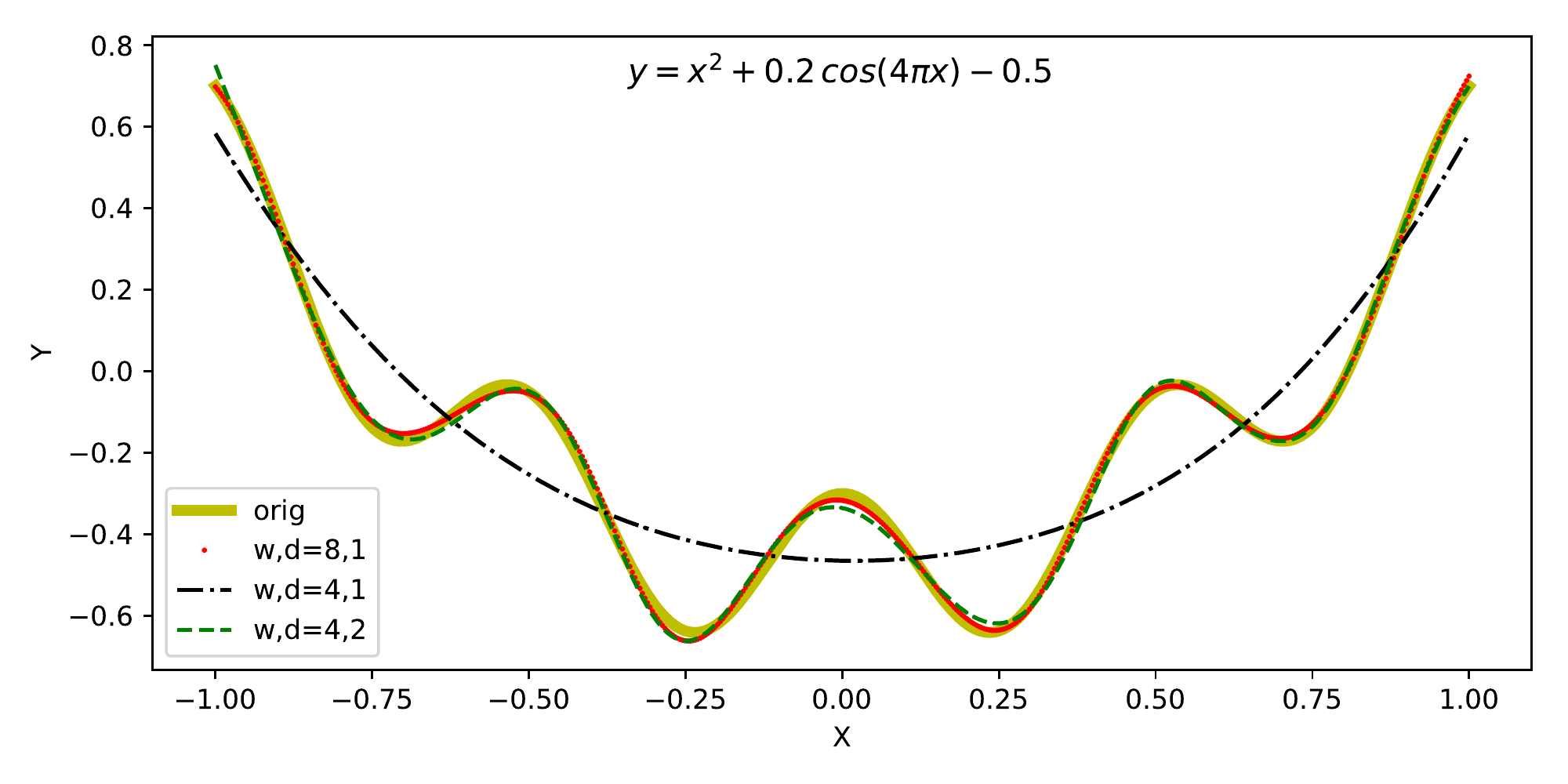}
		\caption{Varying width and depth networks regressing over a dataset with symmetric features}
	\label{fig:examp_symmetric}
\end{figure}

In this section, we take a few examples to distinguish between the additive and multiplicative learning capacity of width and depth, respectively towards assimilating a dataset. We begin by considering a dataset with symmetric features as shown in \textbf{Fig. \ref{fig:examp_symmetric}}. A naive counting would suggest the number of linear features to be $8$. This can be easily seen since the basis is linear prior to \textit{tanh} activation that projects it into a high dimensional space dictated by the number of samples. In other words, there are $8$ slopes that must be accounted for. However, since the features are symmetric we can design two networks that fit the dataset with the same accuracy. 

Fig. \ref{fig:examp_symmetric} shows that Network 1 having $(w,d)=(8,1)$ generates the same accuracy as Network 3 of $(w,d)=(4,2)$. It is important to note that any network of width less than 4 and arbitrary depth will disregard features in this dataset and hence loose accuracy. Network 3 lets us preempt a conclusion that the depth renders multiplicity to the additive features assimilated by the network width. Further, a \textit{sufficient width} is necessary for the depth to become useful in this fashion. The following numerical experiments lead us to a conjecture that the sufficient width has an acute dependence on the number of asymmetric (without multiplicity) features in the dataset and not solely data dimensionality as suggested by a number of Universal Approximation Theorems (UAT) \citep{hanin2017approximating, lu2017expressive}. In this work, we only operate in $L_{2}$ space however similar examples can also be constructed for $L_{p}$ spaces.

We take two examples to substantiate the aforementioned arguments and introduce a notion of \textit{basis collapse} as an obstacle in designing a low-weights, interpretable dense network. The two datasets are generated synthetically using symmetric and asymmetric functions in the absence of noise. The choice of the functions let's us count the number of additive and multiplicative features that a network must assimilate for a good fit. The conclusions reached apply equally to higher dimensional datasets, as explained in the later sections. However, these one-dimensional datasets help us to easily visualize and communicate a few key points of network design. In the following examples, the network depth is held at $2$ considering only width variation.

\subsection{Network Width} \label{ssec:net_width}
\begin{figure}[h]
    \centering
    \begin{subfigure}[b]{\linewidth}
		\centering
		\includegraphics[width=\linewidth]{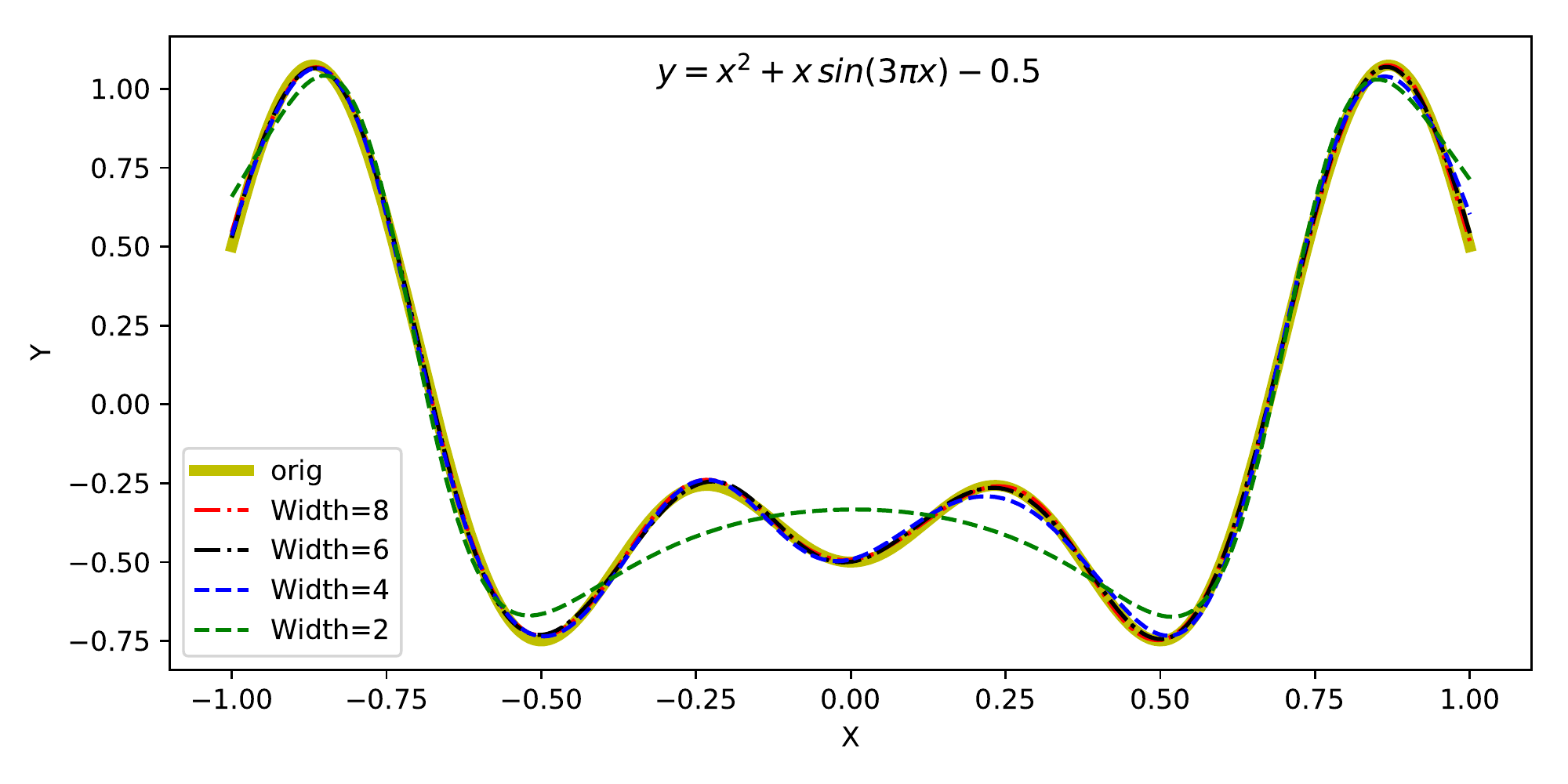}
		\caption{Symmetric}
	\end{subfigure}
	\begin{subfigure}[b]{\linewidth}
		\centering
		\includegraphics[width=\linewidth]{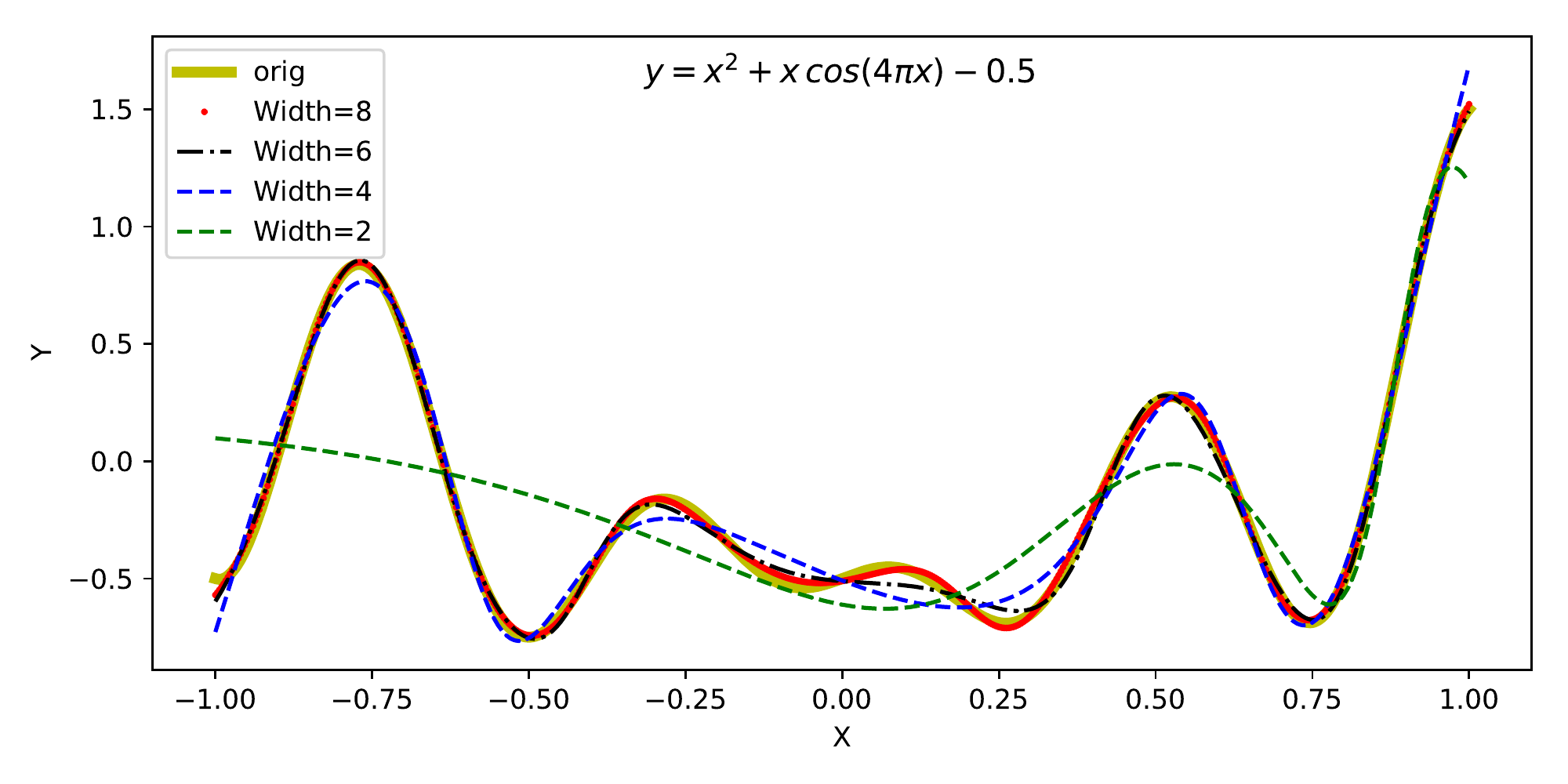}
		\caption{Asymmetric}
	\end{subfigure}
    \caption{Width Variation}
    \label{fig:width}
\end{figure}
\textbf{Fig. \ref{fig:width} (a)} and \textbf{(b)} show the fits corresponding to symmetric and asymmetric functions, respectively with varying widths. It is easy to see that for the symmetric case, a minimum network width of $4$ is sufficient to account for all the features due to multiplicity in the depth. Similarly, a width of $9$ is necessary for the asymmetric function to sufficiently approximate (lower loss value) the $9$ features. Any width less than this sufficient number will not account for the additive features and requires an exponential depth to compensate.

\subsection{Network Depth}
\begin{figure}[h]
    \centering
    \begin{subfigure}[b]{\linewidth}
		\centering
		\includegraphics[width=\linewidth]{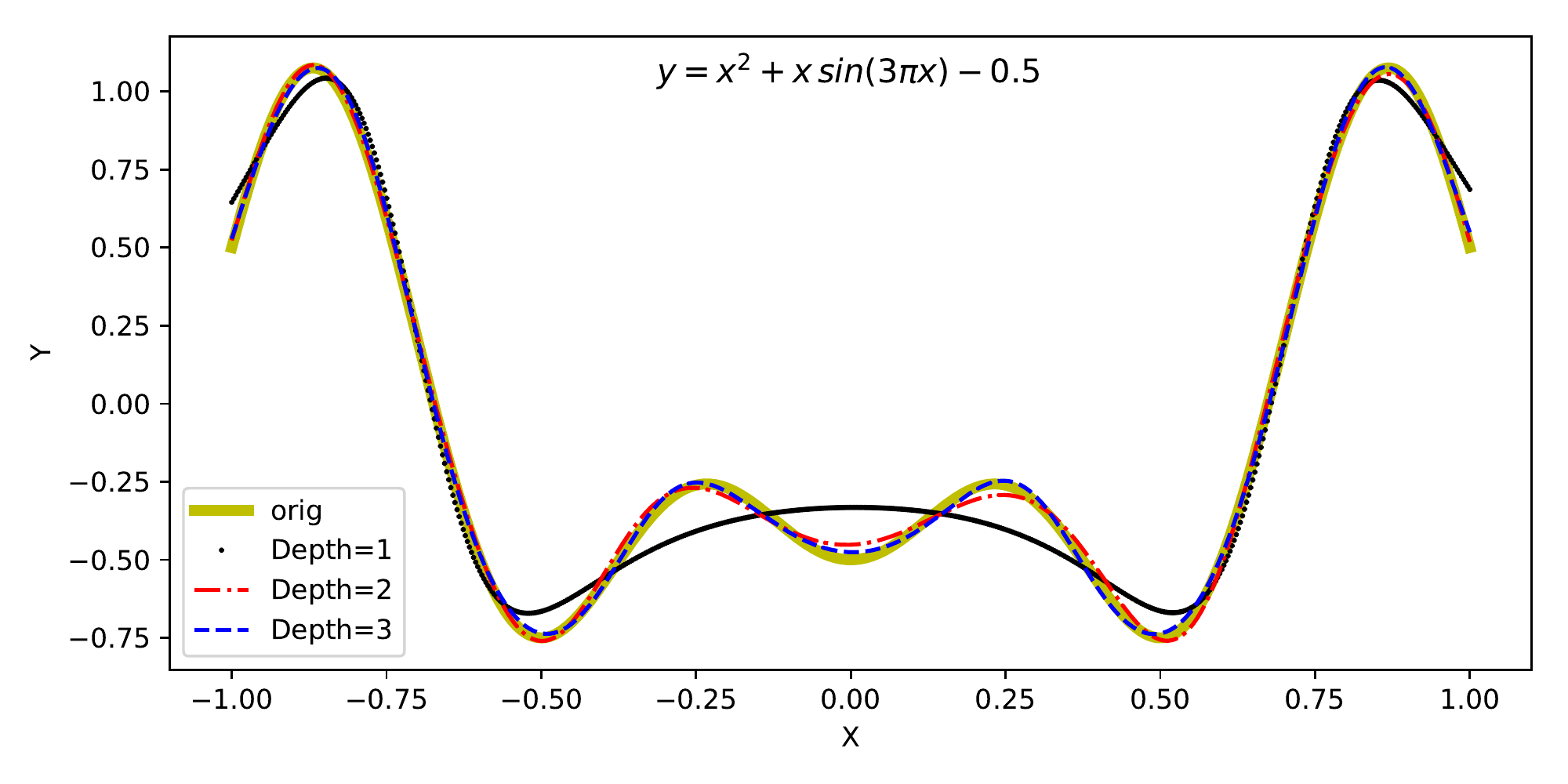}
		\caption{Symmetric}
	\end{subfigure}
	\begin{subfigure}[b]{\linewidth}
		\centering
		\includegraphics[width=\linewidth]{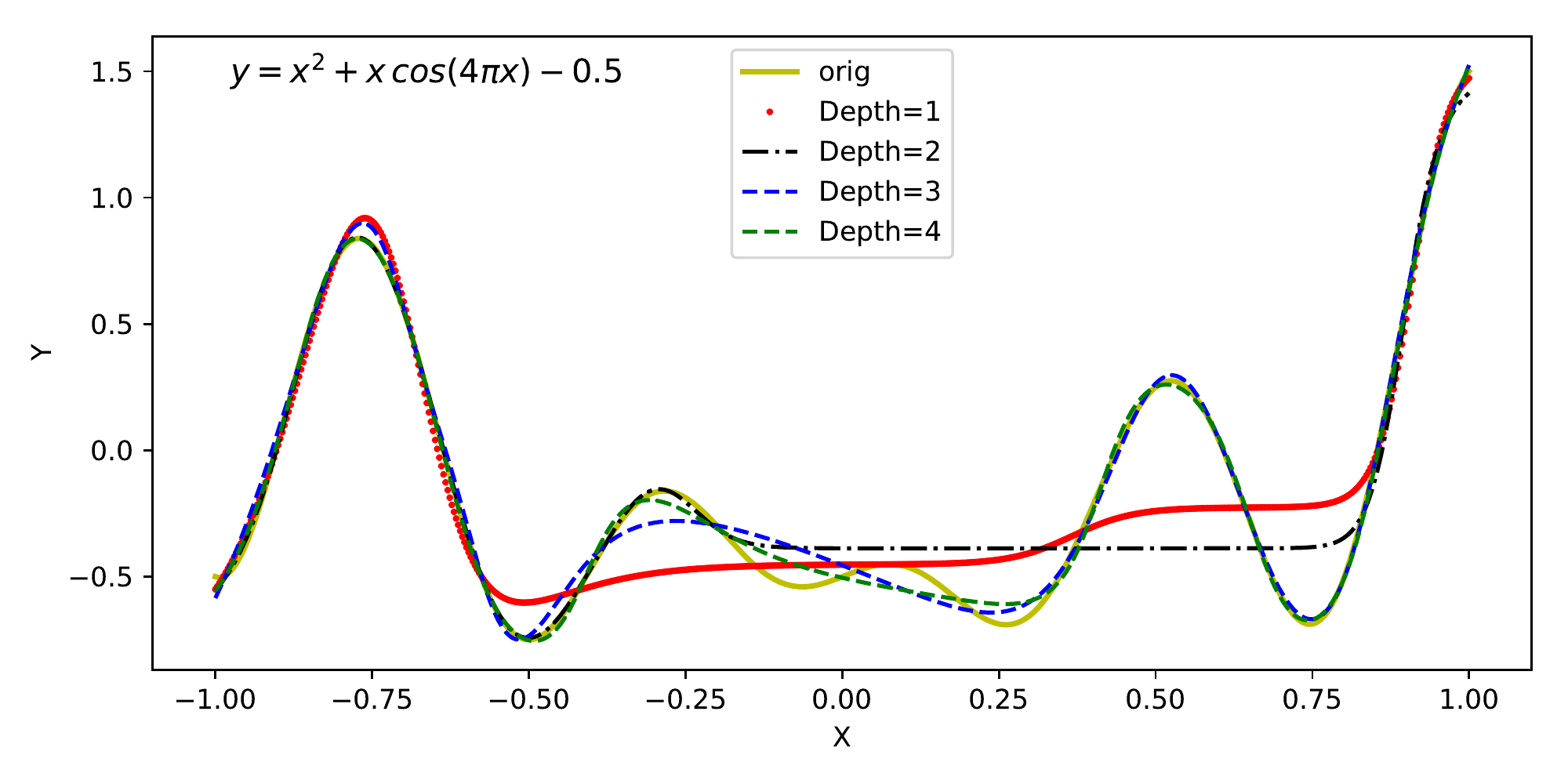}
		\caption{Asymmetric}
	\end{subfigure}
    \caption{Depth Variation}
    \label{fig:depth}
\end{figure}

Similar to the previous subsection, we now consider depth variation while the width is held at $4$. We do not show the cases where the network width is held at $8$ since we are already aware that this can trivially fit the dataset with depth $1$ from the above subsection. \textbf{Fig. \ref{fig:depth} (a)} and \textbf{(b)} show the fits generated using symmetric and asymmetric functions, respectively. It can be easily seen that any depth $(d)$ greater than $2$ will account for the multiplicity of the additive features (width $(w)$) in the network depth for the symmetric dataset. However, for the asymmetric case a width of $4$ is not sufficient and does not account for all the additive features. An increasing network depth now attempts to compensate, at the cost of increasing number of parameters to approximate the features. This in turn results in an overfitting, as shown in Fig \ref{fig:depth} (b), of missing additive features with increasing depth to reduce the loss value. A strange consequence of this is a loss of generalization over narrow domains with missing features.

Although not shown here, we would like to report that even with $(w,d)=(4,10)$, the network converges to a higher loss value when compared to a network with $(w,d)=(8,1)$. 
Also note that in using the term approximation we mean that there is a finite error (although low) in approximating any function even when sufficient finite width is prescribed. If the task at hand is classification, distinct features can still be accounted for with sufficient width. However, for a regression task the requirement of reducing this approximation error is more strict.

\section{Basis Collapse}

After the above array of trivial, nevertheless useful, numerical experiments we would now like to report an elephant in the room with two questions. Given that we already know the exact function from which the dataset was generated: (1) For all the above numerical experiments, how many times we had to train a network from scratch to achieve a good fit? (2) Why does a sufficient width and depth network sometimes fail to account for all the features? 

\begin{figure}[h]
    \centering
    \includegraphics[width=\linewidth]{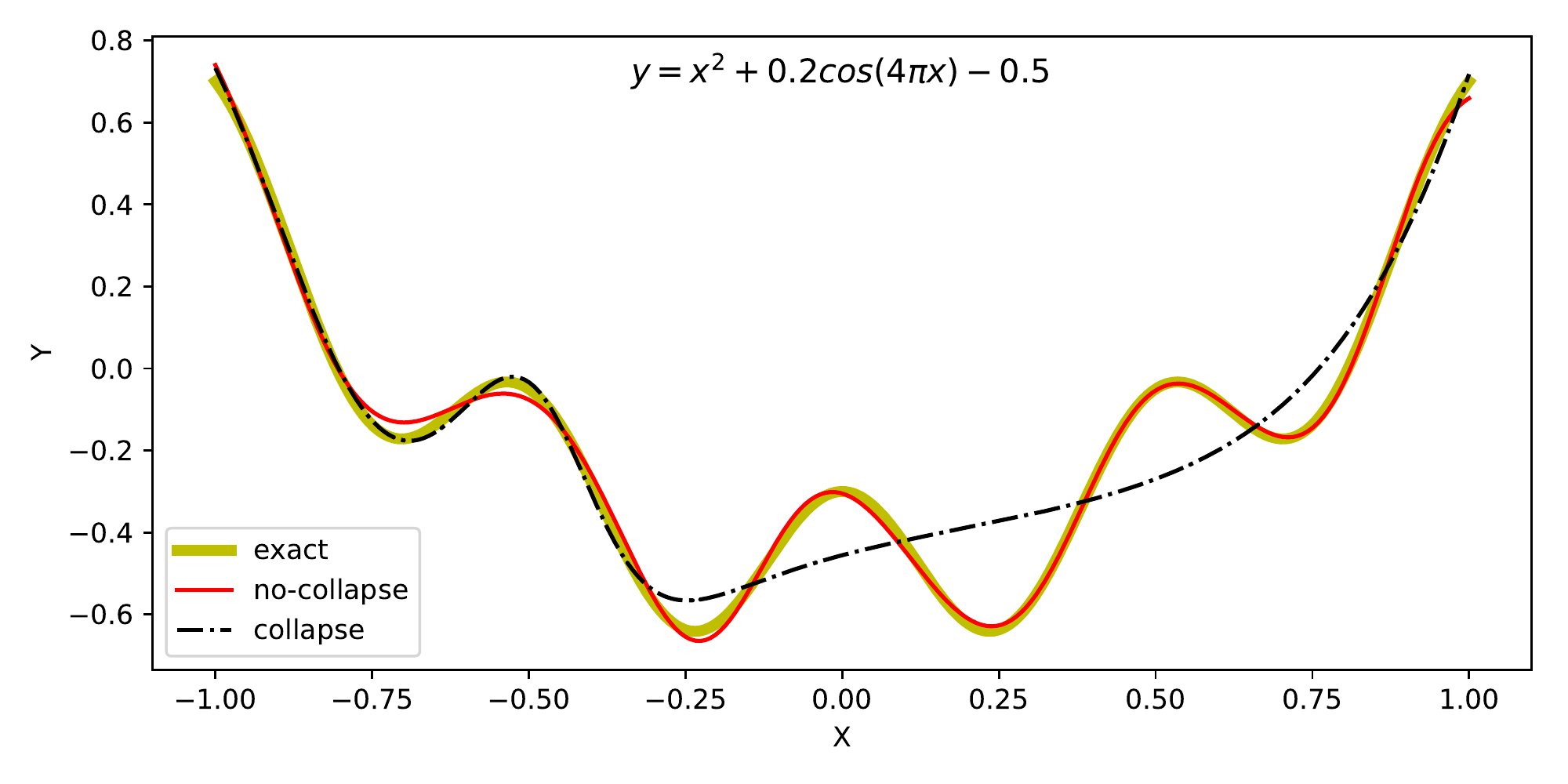}
    \caption{Comparison of collapsed and un-collapsed bases outcome for width 8 and depth 1 network over two runs.}
    \label{fig:collapse}
\end{figure}

The answer to the first question is $>>1$ depending on the number of asymmetric features in the datset. If computational efficiency is a matter of concern this issue is a deal breaker wherein multiple runs must be performed to hopefully identify one desired solution. The second answer is an obstacle that we call \textit{basis collapse} and is the culprit behind both the aforementioned questions. Let us consider the symmetric dataset in Fig, \ref{fig:examp_symmetric} with 4 asymmetric (additive) features. As shown before, both a $(w,d)=(8,1)$ network (Network 1) as well as a $(w,d)=(4,2)$ network (Network 3) can sufficiently account for all the features. We begin by considering only Network 1 to clarify the notion of this basis collapse. \textbf{Fig. \ref{fig:collapse}} shows the fits attained by training the same network multiple times. 

Ideally, a good fit ($mse \leq 10^{-4}$) is achieved if each of the 8 additive features learned in the width are distinct.  However, the network is also prone to replicating a single additive feature more than once (let us say $0<k\leq 8$ times). The consequence now is that we get an $8-k+1$ feature dimensional approximation to the dataset even though a global minimum does exist that corresponds to the desired 8 feature dimensional approximation. One can say an easy fix is adding a second layer $(d=2)$ to the same network. However, note that the lower bound on the width is dictated by $4$ asymmetric features for this dataset. This implies that even when increasing the depth to $2$ or more, the first layer must at least account for $4$ asymmetric features for the depth to become useful as discussed in Sec.  \ref{ssec:net_width} above. 

These findings can be easily verified by performing a singular value decomposition on the layer outputs to identify the redundant or replicated features. Alternatively, for the above simple numerical experiments one can compare the weights and biases of the linear basis (prior to composing with activation) in a single layer. In conclusion, given sufficient width and depth the network must still learn all the asymmetric additive features in the first layer for the network to arrive at a good fit. Since we have the luxury of knowing the exact function we can identify when this happens. However, to a network designer working with an arbitrary high-dimensional dataset this luxury is not affordable. The network loss not achieving lower values then gives an impression that the network width and depth are not sufficient and must therefore be increased to achieve a better loss value. We now provide two solutions to this problem, towards designing a low-weights network, that under the arguments of additive and multiplicative learning renders interpretability. 

\noindent \textbf{(1) Band-aid Solution (BS):} Increase the network width and depth until you achieve a desired loss or fit followed by SVD on each of the layer 1 outputs/features to identify redundancy. Reducing the network width using this approach will still ensure that the $m$ dimensional feature solution achieved by the larger network is still available to the smaller network. Redundancy in the network depth does not pose any problems in interpretation and in fact contributes to a well conditioned minimization problem as discussed in the \ref{ssec:scale_independent} above. 



\noindent \textbf{(2) Loss Enforced Bases (LEB):} The above approach is tedious (from authors own experience) and requires post-processing to prevent basis collapse. A more general solution is to directly inform the loss with a similarity metric that the layer 1 outputs must represent distinct features. We prefer this approach since it is more general and requires only one tuning parameter. The similarity metric ($loss_{sim}$) is a projection of the Euclidean distances between the features into a feature dimensional Gaussian space given by,
\begin{equation}
    loss_{sim} = exp\left({-\frac{1}{2N\sigma }\sum_{i,j,i\neq j}^{w}||l_{1,i}-l_{1,j}||_{2}^{2}}\right).
\end{equation}
Here, $w$ is the network width, $l_{1,i}$ is the $i^{th}$ output feature vector of layer 1, $N$ is the length of the feature vector, and $\sigma$ is the only tuning parameter to adjust the distance between the output feature vectors. It can be easily seen that $loss_{sim}$ is bounded above and below by 1 and 0. This additional similarity loss function can be scaled in accord with the choice of activation/localization function as discussed before in \ref{ssec:scale_independent}. Since we are mainly relying upon \textit{tanh} for our numerical results, this form is sufficient for our choice of $L_{2}$ loss between the true and predicted labels for any dataset. A tuning parameter ($\sigma$) value of 0.01 to 0.001 worked well for all of our numerical experiments. 

This additional similarity loss along with \textit{mse} loss ensures that we circumvent basis collapse issues that brings the loss drop rate to a screeching halt. In doing so, we also avoid the need for Dropout \citep{srivastava2014dropout} as a means to prevent overfitting while keeping the network design low weights. As shown in Fig. \ref{fig:depth} (b), increasing network depth in the event of a basis collapse will only lead to overfitting of the missing additive features. Our approach mitigates this issue while ensuring that every feature learned by the network width remains distinct (not to be confused with orthogonality) by enforcing a bounded similarity loss. We can now be confident in our expectation that as the network width increases we will obtain a better loss. An additional \textit{positive outcome} is that we consistently get an approximately similar fit even when multiple training runs are prescribed in our numerical experiments.

\section{Universal Approximation Theorems}

\begin{figure}[h]
    \centering
    \begin{subfigure}[b]{0.49\linewidth}
		\centering
		\includegraphics[width=\linewidth]{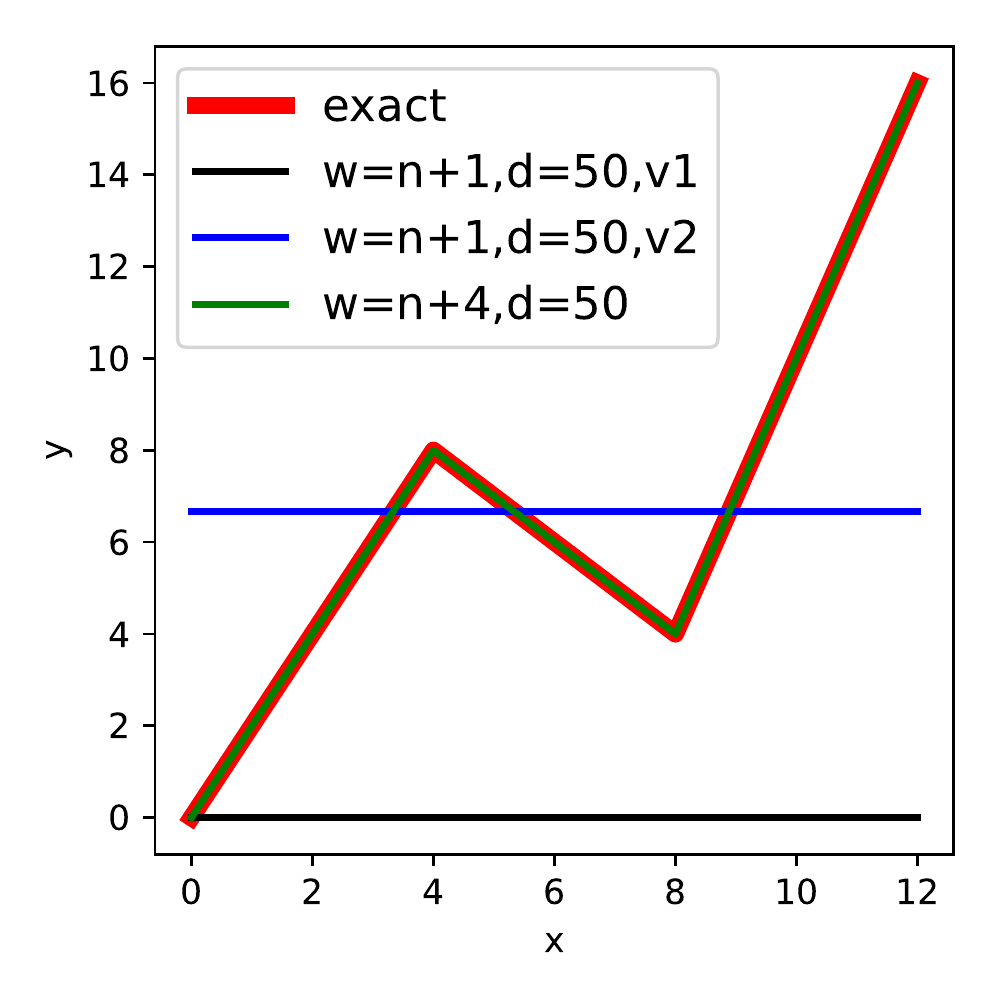}
		\caption{Predicting $y=5x$}
	\end{subfigure}
	\begin{subfigure}[b]{0.49\linewidth}
		\centering
		\includegraphics[width=\linewidth]{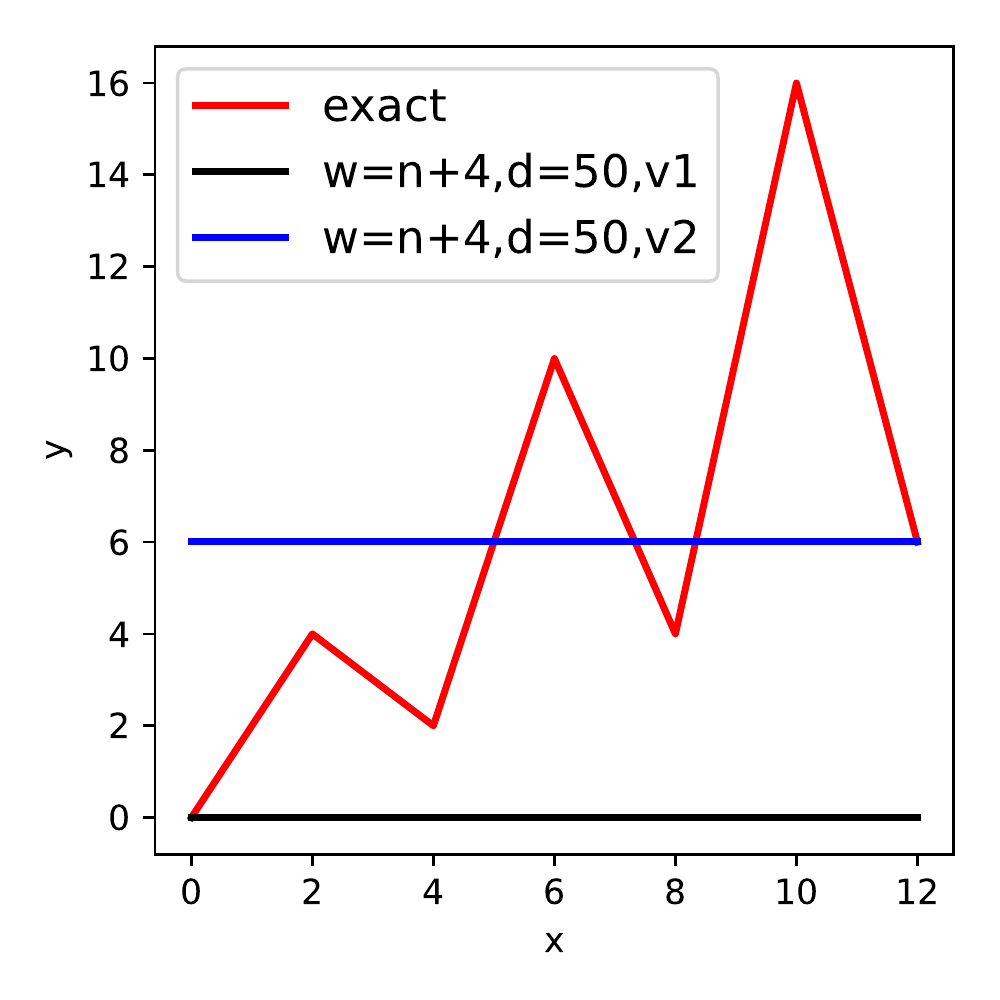}
		\caption{Predicting $y=5x^2$}
	\end{subfigure}
    \caption{Validating the claims of UAT where width is guided by the input dimension $\mathbb{R}^n$. Although it seems that $w=n+4=5$ net for $1D$ data works for (a) because it has three features, it immediately fails in (b), where the input features are six.}
    \label{fig:uat}
\end{figure}

Recent theoretical works on UATs states that for a given width, if depth tends to infinity, then any continuous function can be approximated. To mention a few, Hanin \etal \cite{hanin2017approximating} and Lu \etal \cite{lu2017expressive} state that if the input dimension is $n$, then any function in $\mathbb{R}^n$ can be approximated by $n+1$ and $n+4$ width respectively, using \textit{relu} activation. However, our numerical experiments contradict these findings and suggest that there is a missing component of feature complexity that must be taken into account for generalization. In the following, we attempt to reproduce these findings using simple numerical experiments.

To confirm these findings, we simulated data as $L_{1}$ functions in 1-D domain with a Mean Absolute Error ($mae$) as per the description in the references above. In \textbf{Fig. \ref{fig:uat} (a)}, since the $1D$ data has three distinct linear features, a network of width $3$ should be adequate. We replicated the setup following \cite{hanin2017approximating} with width $w=n+1=2$ neurons at each depth for varying depths over multiple runs. Fig. \ref{fig:uat} shows that the network either converged to a fully collapsed (zeros) or partially collapsed (data mean) state. 

On the other hand, the setup in \cite{lu2017expressive} suggests that a network of width  $w=n+4=5$ universally approximates this function, since only three features $w=3$ is sufficient for this curve. However, this result was achieved in $\mathbf{1}$ out of $\mathbf{113}$ runs of the same network, even while using similarity loss (with $L_{1}$ distances) as discussed before. Fig. \ref{fig:uat} (a) shows two such collapsed cases as v1 and v2. This is attributed to the surface roughness imparted by \textit{relu} activation as described in Sec. \ref{sec:act}.

\textbf{Fig. \ref{fig:uat} (b)} shows another dataset, where the number of distinct linear features are now six in $\mathbb{R}^1$. We find that the argument of $w=n+4=5$ in Lu \etal \cite{lu2017expressive} does not hold, as the numerical experiments either collapse to zero or to the data mean over various runs. However, our arguments of additive and multiplicative features still hold true. The only difficulty here is the challenge posed by \textit{relu} activation and consequently the loss induced hidden parameter surface roughness leading to frequent basis collapse.

\section{Higher Dimensional Datasets}

\begin{figure}[h]
\centering
    \begin{subfigure}[b]{0.49\linewidth}
		\centering
		\includegraphics[trim={2cm 0 1cm 0},clip,width=\linewidth]{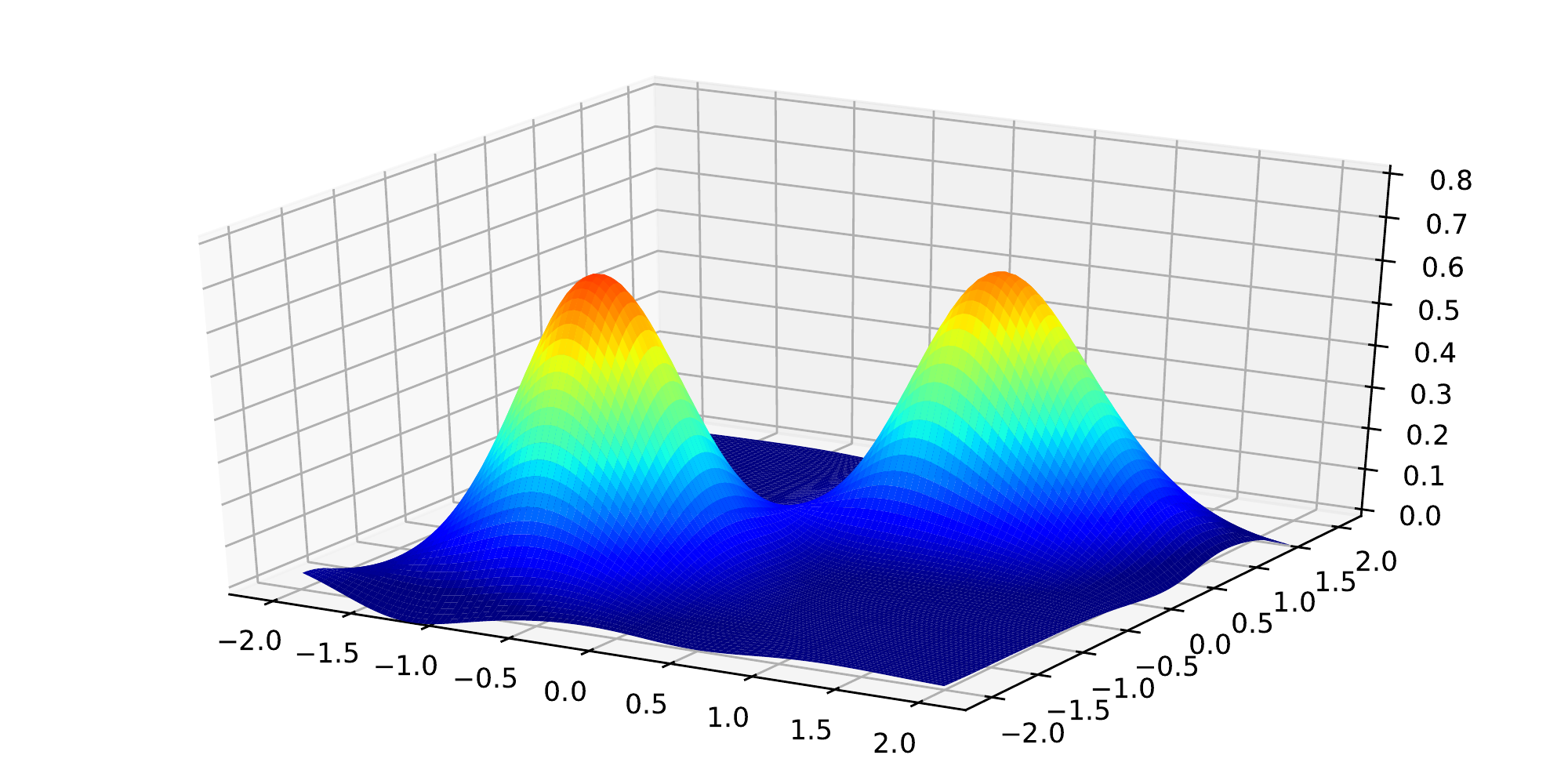}
		\caption{Surface at t = 0.0}
	\end{subfigure}
	\begin{subfigure}[b]{0.49\linewidth}
		\centering
		\includegraphics[trim={2cm 0 1cm 0},clip,width=\linewidth]{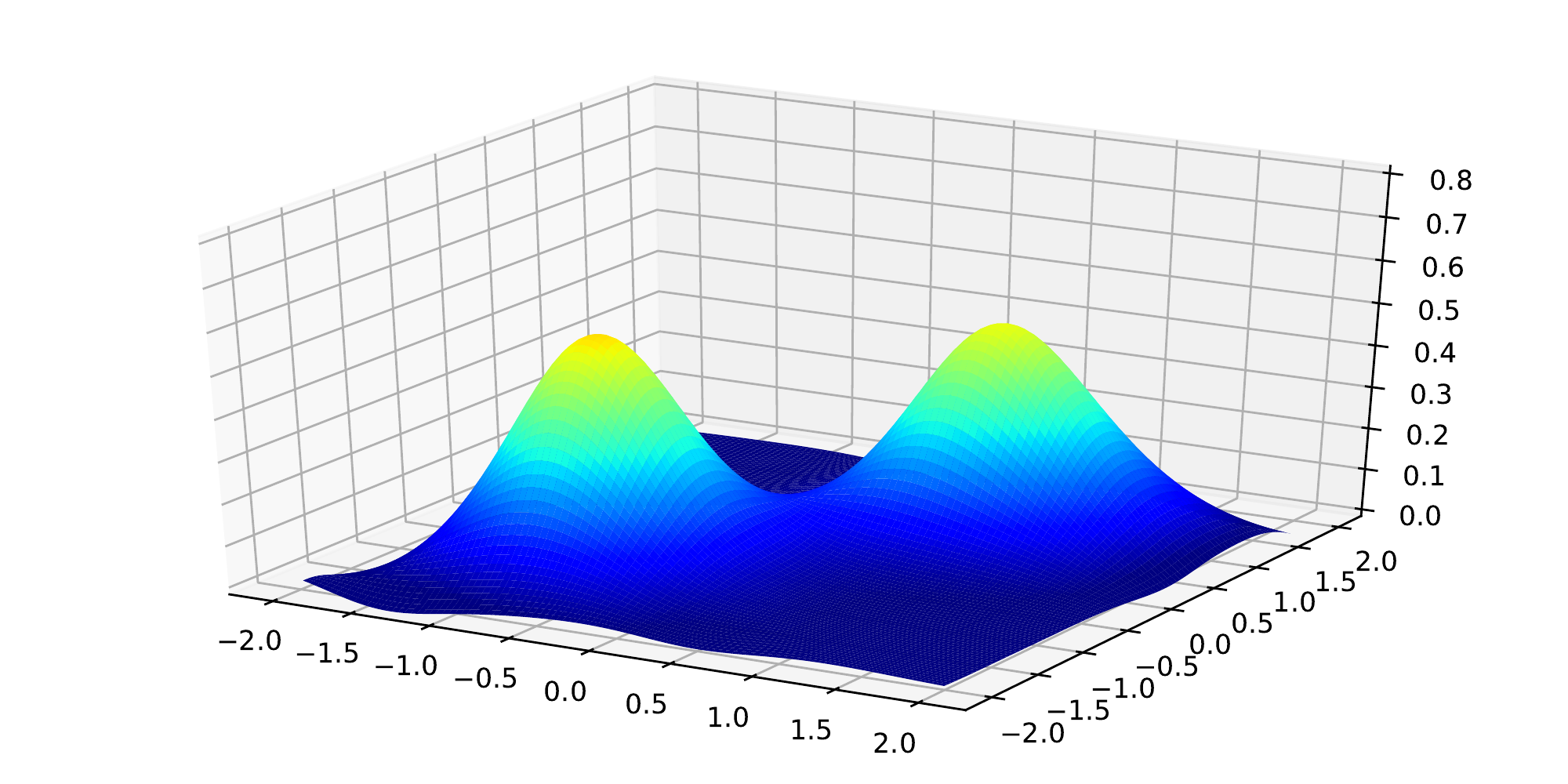}
		\caption{Surface at t = 0.5}
	\end{subfigure}
    \caption{Solving a 3-dimensional dataset with symmetric features. \textit{mse} loss: $3.24e-4$ at $30$ epochs with $lr=0.2$}
    \label{fig:par_2d_sym}
\end{figure}

\begin{figure}[h]
    \centering
    \begin{subfigure}[b]{0.49\linewidth}
		\centering
		\includegraphics[trim={2.5cm 0 1cm 0},clip,width=\linewidth]{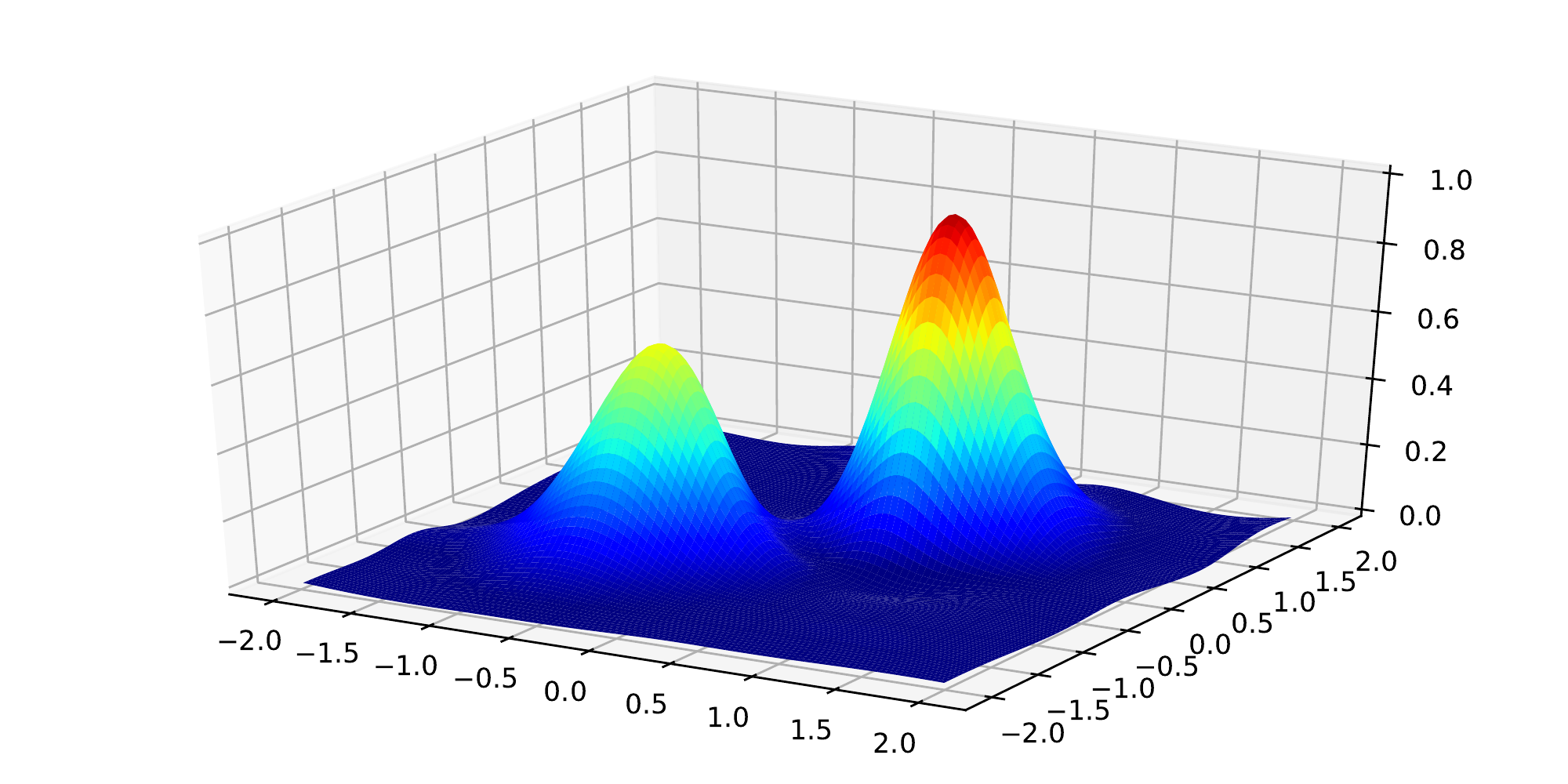}
		\caption{Surface at t = 0.0}
	\end{subfigure}
	\begin{subfigure}[b]{0.49\linewidth}
		\centering
		\includegraphics[trim={2.5cm 0 1cm 0},clip,width=\linewidth]{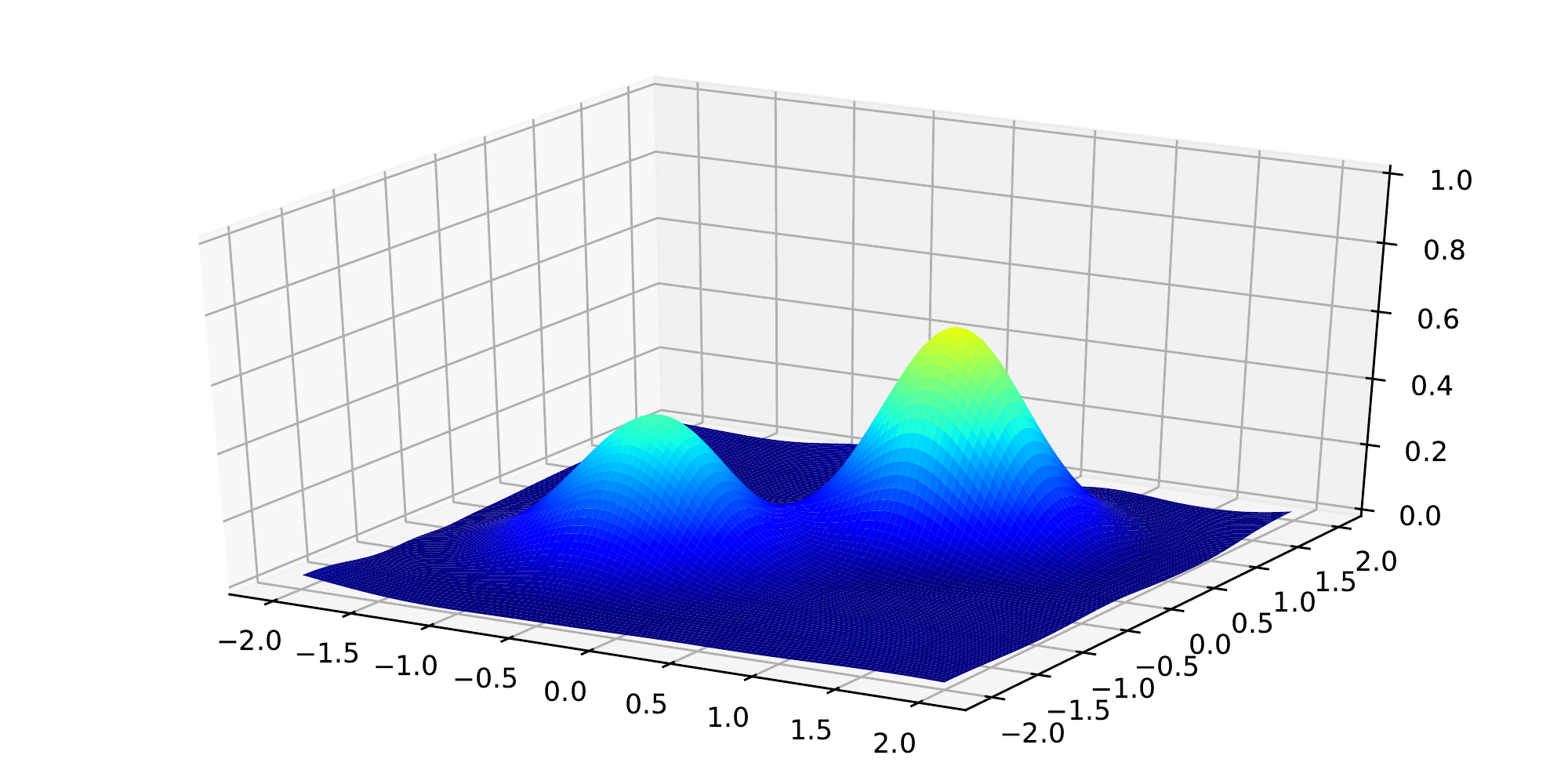}
		\caption{Surface at t = 0.5}
	\end{subfigure}
    \caption{Solving a 3-dimensional dataset with asymmetric features. \textit{mse} loss: $5.81e-4$ at $30$ epochs with $lr=0.2$}
    \label{fig:par_2d_asym}
\end{figure}

With the \textit{basis collapse} issue sufficiently addressed, we are now ready to demonstrate that the aforementioned arguments hold for higher dimensional datasets. We present two, $3D$ datasets $(X \times Y \times T)$ with symmetric and asymmetric features as shown in Figs. \ref{fig:par_2d_sym} and \ref{fig:par_2d_asym}, respectively. The symmetric dataset was regressed using a $(w,d)=(4,4)$ network whereas the asymmetric dataset requires a $(w,d)=(8,4)$ network to converge to a loss at scale $1e-4$ in both the cases.

The dataset is a temperature profile of a surface as heat dissipates over time. In order to generate symmetric features we consider a surface with two spherical objects, at constant and equal temperature, placed on top of this surface, as the initial state. This is shown in \textbf{Fig. \ref{fig:par_2d_sym} (a)}. Once the spherical objects are removed the heat dissipates generating a temperature profile over time. Similarly, for the asymmetric case the two spherical objects, at constant but unequal temperatures, are considered as shown in \textbf{Fig. \ref{fig:par_2d_asym} (a)}. Each of the datasets consist of two spatial and one temporal dimensions.

It is not at all surprising that the network size for the symmetric case in this section is the same as the one used to generate Figs. \ref{fig:bg_2d_f1} and \ref{fig:bg_2d_f2}. In fact, now it is clear that depth can render multiplicity to the additive features learned in the width to account for symmetries. It is our understanding that as the data dimensionality and feature complexity increases, visualization becomes a limited tool to draw such conclusions to reach a low weight network. However, these visually viable examples are instructive in designing robust and transparent dense nets even in cases where visualization is of limited avail.

\section{Data \vs Representation Driven}

In this section, we discuss the differences between conventional data-driven and recently introduced representation-driven by Raissi \cite{raissi2018deep} driven dense networks. The core objective of a data-driven network is to regress to a representation purely from data. The classic ImageNet \cite{deng2009imagenet} challenge in Computer Vision is to classify images into classes with the underlying network extracting a representation. If the true manifold (solution as a function of the input images) were already known, the field would culminate in a complete success. On the other hand, the Physics Inspired Neural Network (PINN) proposed by Raissi \etal \cite{raissi2019physics} relies upon supplying a known representation as a loss function to generate the solution surface. This is to say that the manifold is either already known to us as a function prescribed as an Ordinary or Partial Differential Equation (ODE, PDE) with accompanying initial and boundary conditions. 

Under our description of feature learning in width and depth, we show that an appropriately designed dense network remains indifferent to this choice of data or representation driven forms. Further, the network size remains unchanged if the discrete solution surface is shown as training labels or as a representation informed loss function while achieving comparable loss values at scale. To demonstrate this equivalence, we begin with three trivial examples to mimic exact, ODEs or PDEs and later apply our core guidelines to the PINN framework to construct equivalent low-weights networks with $\mathbf{100\times}$ fewer parameters and $\mathbf{10\times}$ lower MSE.

\begin{table}[h]
    \centering
    \begin{tabular}{|c|c|c|c|c|c|c|} \hline
        Form & Regress & Exact & ODEv1 & ODEv2 & PDEv1 & PDEv2 \\ \hline
        MSE (e-4) & $1.85$ & $2.01$ & $2.03$ & $2.11$ & $4.76$ & $4.94$ \\ \hline
    \end{tabular}
    \caption{Comparison of Data \vs Representation Driven. Dense Net of $(w,d)=(4,2)$, Epochs: 30, $lr=0.2$}
    \label{tab:comp_forms}
\end{table}

In the light of our simple numerical experiments in Sec. \ref{sec:extrap}, the extrapolation capacity of a dense net outside the trained domain is questionable. This is true for all such nets that we trained on $3$D physical datasets considered  throughout this paper. This is to say that a dense net trained for a representation given one set of initial and boundary conditions, certainly cannot be used to extract a different representation with another set of initial and boundary conditions. Hypothetically, to achieve the above, a network must be trained on both sets. In fact in order to represent an ODE/PDE, the network must be trained on a family of initial and boundary functions. 


We again start with a few simple numerical experiments using a dense net of $(w,d)=(4,2)$ with \textit{tanh} activation to build our understanding. A known representation (parabola) is prescribed using 1. functional, 2. ODE and 3. PDE forms, with \textit{mse} reported in \textbf{Table. \ref{tab:comp_forms}}. This is followed by a comparative study against the PINN framework \cite{raissi2018deep} wherein a careful curation of PDE representations done by the authors helps us to show the effectiveness of our low-weights framework.

\subsection{Exact Equation}
\begin{figure}[h]
    \centering
    \begin{subfigure}[b]{0.49\linewidth}
		\centering
		\includegraphics[width=\linewidth]{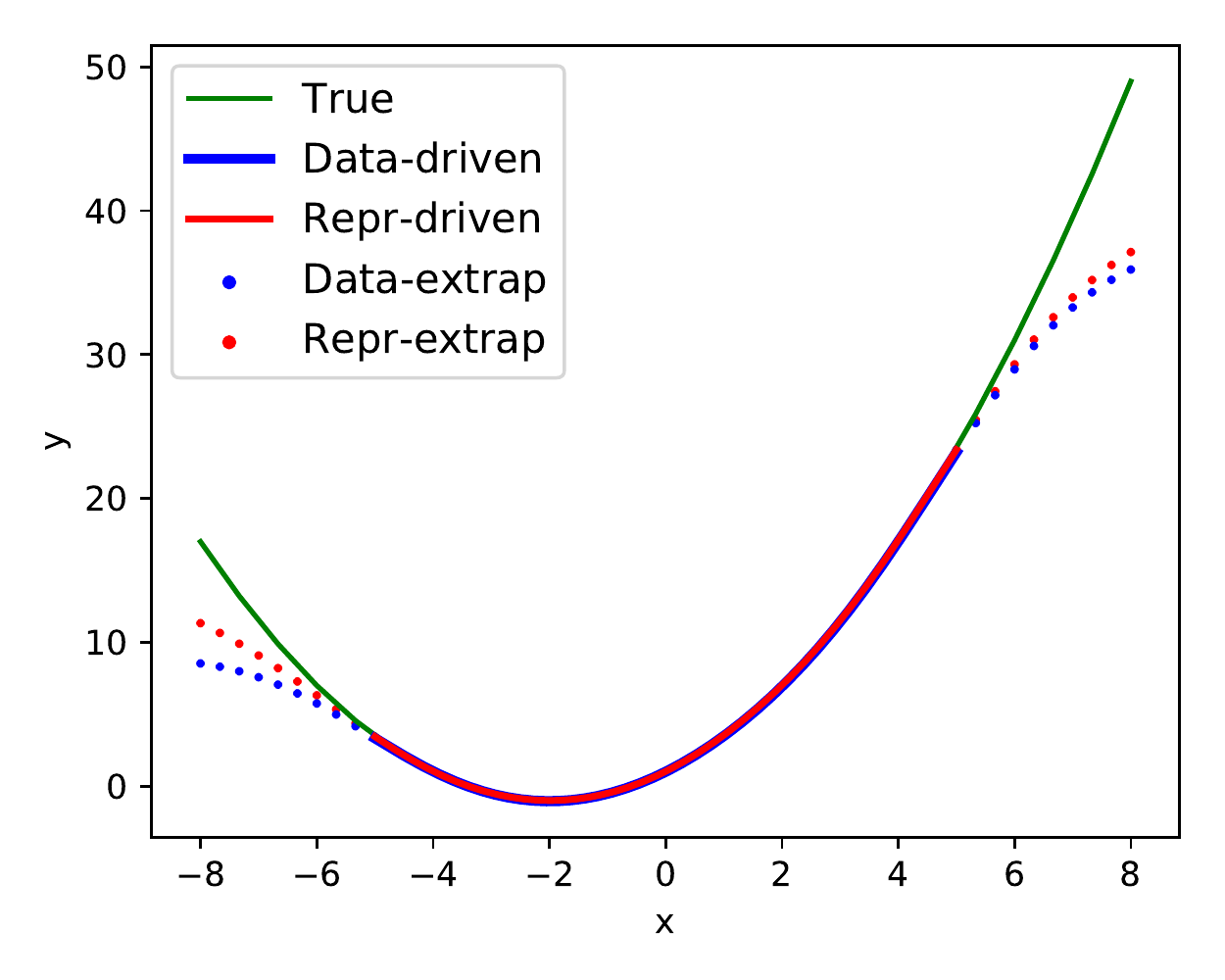}
		\caption{Exact Form}
	\end{subfigure}
	\begin{subfigure}[b]{0.49\linewidth}
		\centering
		\includegraphics[width=\linewidth]{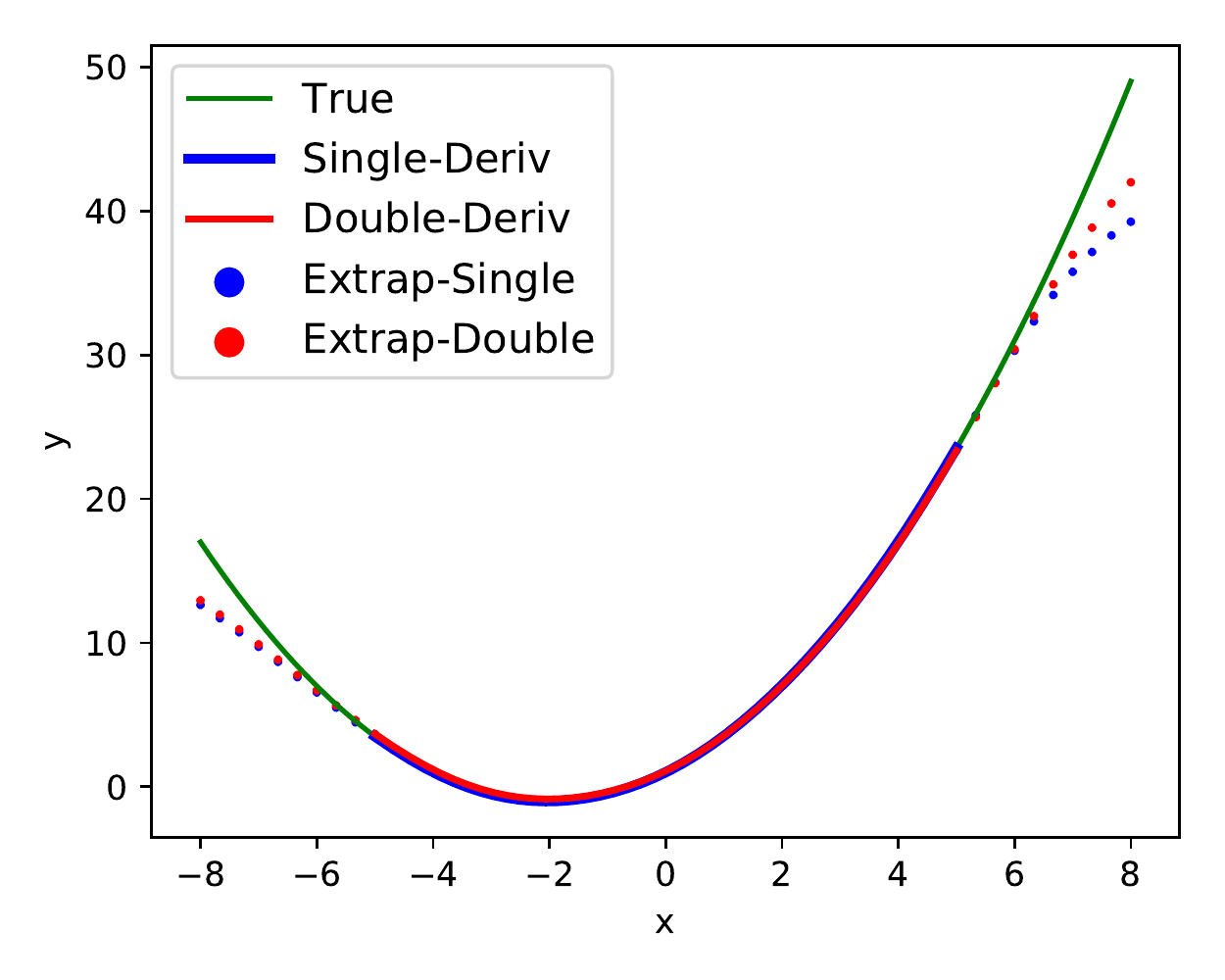}
		\caption{ODE Form}
	\end{subfigure}
	\caption{Fitting a functional representation prescribed by an Exact form and ODE with initial condition.}
    \label{fig:exactpara}
\end{figure}

 We first consider a parabolic equation $y = f(x) = 0.5x^2 + 2x+ 1$ in the domain $x=[-5,5]$, which resembles a shifted parabola shown in \textbf{Fig. \ref{fig:exactpara}}. The data driven network is supplied values of $x$ at input and learns to regress to its corresponding $y$ at output, via \textit{mse} loss. For a representation driven network, we explicitly specify the loss as  $\frac{1}{N}||(y_{pred}-f(x))||^{2}_{2}$ where N is the number of samples.
Fig. \ref{fig:exactpara} (a) shows that both data and representation driven examples provide an adequate fit. However, the extrapolation deviates from the exact form for both data and representation driven dense nets.


\subsection{Ordinary Differential Equation (ODE)}

We now consider an equivalent ODE solution where initial condition is prescribed such that the solution of the form  $y = f(x) = 0.5x^2 + 2x+ 1$. The first order ODE is then $y_{x} = 2(0.5)x + 2$ with $y(0)=1$ as the initial condition. Similarly, a second order ODE can be written as $y_{xx} = 2(0.5)$ with intermediate condition $y_{x}=2(0.5)x+2$, and initial condition $y(0)=1$. Here the loss functions for the first and second order forms are augmented with the additional conditions as $||y_{x}=2(0.5)x+2||_{2}^{2}$ and $||y(0)-1||_{2}^{2}$, respectively.
As can be seen in Fig. \ref{fig:exactpara} (b), both prescriptions of loss functions adequately fit the data inside the trained domain. The extrapolation is accurate for a narrow region outside the domain and then deviates quickly by inhereting the smoothness of \textit{tanh} activation as expected.

\subsection{Partial Differential Equations (PDE)}
\begin{figure}[h]
    \centering
    \includegraphics[width=0.6\linewidth]{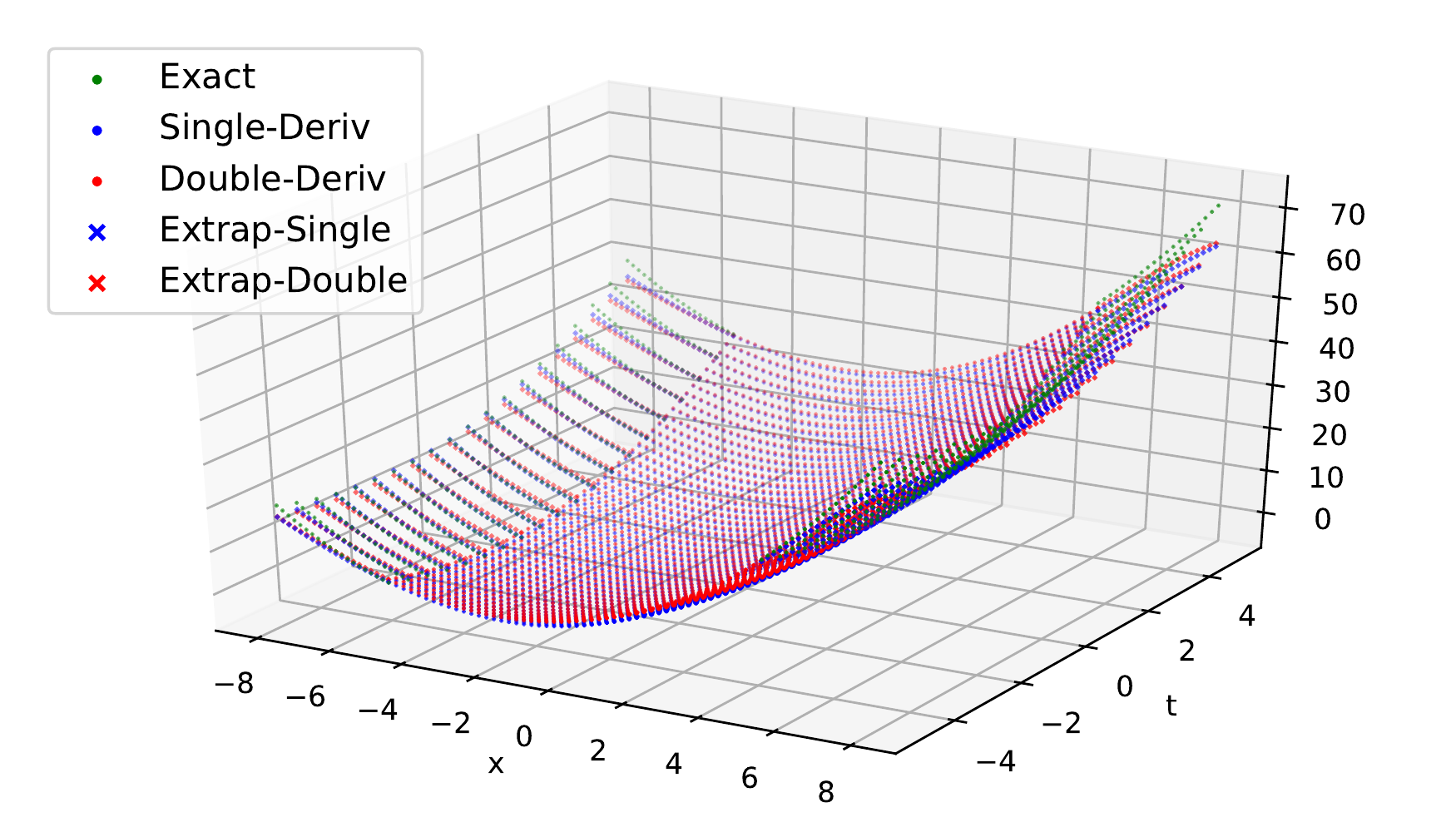}
	\caption{Fitting a representation prescribed by a PDE with accompanying manifold conditions on the boundaries.}
    \label{fig:pde}
\end{figure}

Finally in \textbf{Fig. \ref{fig:pde}}, we consider a multivariate function in two variables $y = f(x,t) = 0.5x^2 + 2x+ 0.5t^2 + 2t + 1$ with an equivalent PDE form $y_{x} + y_{t} = 2(0.5x+0.5t) + (2+2)$ with the initial condition $y(x,-4) = 0.5x^2+2x+17$, and boundary conditions $y(t,4) = 0.5t^2+2t+17$ and $y(t,-4) = 0.5t^2+2t+1$. On similar lines, a second order PDE form can also be prescribed which we curtail here for the sake of conciseness.
The fitting inside the domain is quite good whereas the extrapolation quickly deviates by inheriting properties of \textit{tanh} activation.

\subsection{A Comparative Study}

We now present a comparison of our low-weight network design against the PINN \cite{raissi2018deep} framework. Since the ground truth representations are known, this provides us with an opportunity to verify our arguments of additive and multiplicative learning. As discussed before the network width and depth were kept the same for both data and representation driven learning here. In the following we borrow representations and datasets from Raissi \etal \cite{raissi2018deep,raissi2019physics} for benchmarking purposes. 

\begin{table}[t]
    \centering
    \caption{Comparison of PINN \vs Our Approach. Epochs 40, Steps per Epoch 2000, $lr:0.2$}
    \begin{tabular}{|l|cccc|cccc|} \hline
     & \multicolumn{4}{c|}{Raissi (*Best)} & \multicolumn{4}{c|}{Ours(Mean)} \\ 
    Form & Depth & Width & MSE & Params & Depth & Width & MSE & Params \\ \hline
    Allen Cahn & $4$ & $200$ & $6.99e-3$ & $121401$ & $3$ & $6$ & $3.64e-4$ & $108$ \\ Burgers' & $9$ & $20$ & $4.78e-3$ & $3441$ &  $3$ & $4$ & $1.28e-4$ & $56$ \\ 
    Korteweg de Vries & $5$ & $50$ & $3.44e-2$ & $10401$ & $3$ & $10$ & $4.96e-4$ & $260$  \\ 
    Kuramoto-Sivashinsky & $5$ & $50$ & $7.63e-2$ & $10401$ & $3$ & $20$ & $1.64e-3$ & $920$ \\ 
    Navier Stokes & $5$ & $200$ & $5.79e-3$ & $161601$ & $3$ & $16$ & $5.82e-4$ & $608$ \\ 
    Schr\"{o}dinger & $5$ & $100$ & $1.97e-3$ & $40801$ & $3$ & $6$ & $1.55e-4$ & $108$ \\ \hline
    \end{tabular}
    \label{table:comp}
\end{table}

The examples shown in Figs. \ref{fig:bg_2d_f1},  \ref{fig:bg_2d_f2}, \ref{fig:par_2d_sym}, and \ref{fig:par_2d_asym} were generated on a two spatial and one temporal domain $(X\times Y \times T)$ of size $(-1,1)\times(-1,1)\times(0,1]$. Further, $8000$ uniformly distributed points in the space-time volume were chosen with an equal number of points on the 2-D manifolds that represent the spatial and temporal boundaries. The results show that a low-weight network can adequately represent the features for these famous problems. Since the basis collapse issue is addressed, increasing width now increases the accuracy of the functional approximation. We restrict our results to minimal/sufficient networks that adequately reproduce the features. 

\begin{figure}[h]
    \centering
    \begin{subfigure}[b]{0.49\linewidth}
		\centering
		\includegraphics[width=.97\linewidth]{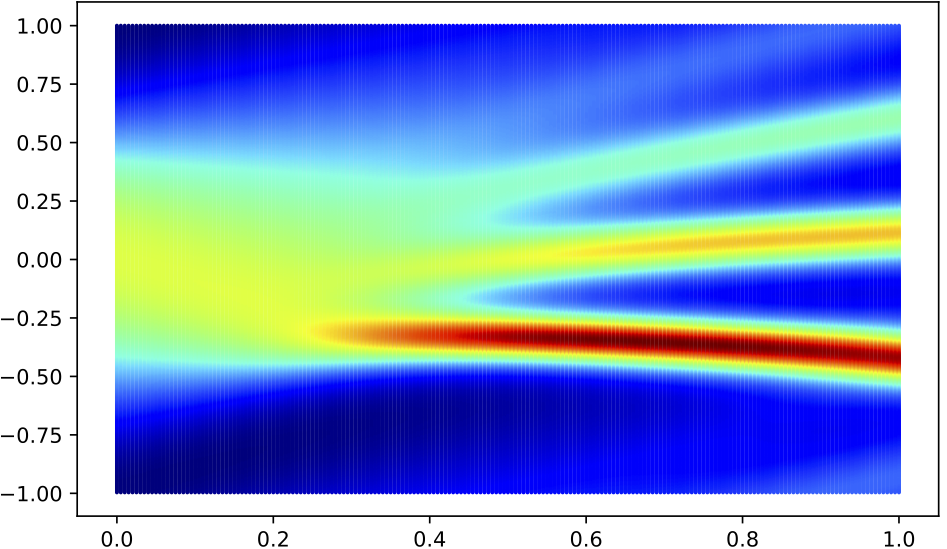}
		\caption{True Solution}
	\end{subfigure}
	\begin{subfigure}[b]{0.49\linewidth}
		\centering
		\includegraphics[width=1.1\linewidth]{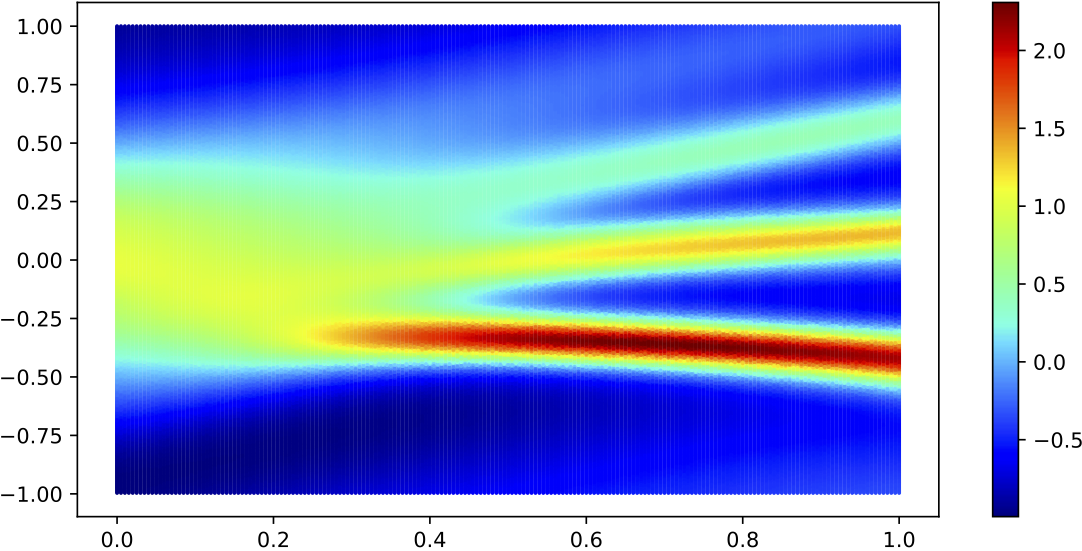}
		\caption{Learned Solution}
	\end{subfigure}
	\caption{Korteweg de Vries Problem}
    \label{fig:kdv}
\end{figure}
As can be seen from \textbf{Table. \ref{table:comp}}, our narrow and shallow networks computationally outperform the wide and deep networks while sufficiently replicating all features. We would also like to report that at the loss values reported here, we visually observe mild, as shown in \textbf{Fig. \ref{fig:kdv}} differences between the exact and learned features even with an $mse$ loss at scale $1.e-4$. Due to limited space, we only show the fitted solution corresponding to the Korteweg de Vries (KdV) form. This is attributed to the function approximation error of our low-weight network. As the network size is increased with increasing point density, we expect, the approximation errors to drop further with at least similar $mse$ values at scale.

For each of the PDEs forms, Allen-Cahn, Burgers', Korteweg de Vries, Kuramoto-Sivashinsky, Navier Stokes and Schr\"{o}dinger, we report $mse$ losses with maximum and minimum training epochs of $30$ and $10$, respectively. The results tabulated in Table. \ref{table:comp}, show that our narrow and shallow networks significantly outperform the previous results in capturing all the features. A comparison of our average MSE values (over multiple runs) against the Best MSE value in Rassi \etal \cite{raissi2018deep} are reported. We achieve these results using a very low weight network by preventing basis collapse and appropriately scaling the output labels/representation in accordance with \textit{tanh} activation.

\section{Conclusion}

We presented transparent design guidelines for low-weight dense nets by addressing a basis collapse issue. Our numerical experiments demonstrate the interpretation of width and depth in terms of learning additive and multiplicative features in datasets. The additional similarity loss term, to prevent basis collapse, is indifferent to the application domain and works well for both data and representation driven frameworks. We verify our network design's feature extraction capabilities on three dimensional datasets with known ground truths. Although the task at hand here was essentially regression we can apply the same concepts to classification problems without any loss of generality. A performance comparison against prior works shows that our design is orders of magnitude ($\sim 100\times$) lesser in number of parameters while achieving lower loss values at scale ($\sim 10 \times$). We also report that preventing basis collapse enhances network reproducibility over multiple training runs with similar features identified in every run.  

\bibliographystyle{unsrt}

\end{document}